\documentclass[sigconf]{acmart}

\usepackage{marvosym}
\usepackage{graphicx}
\usepackage{subfigure}
\usepackage{amsmath}
\usepackage{booktabs}
\renewcommand{\paragraph}[1]{
     \textbf{#1.} 
 }

\let\svthefootnote\thefootnote
\newcommand\freefootnote[1]{%
  \let\thefootnote\relax%
  \footnotetext{#1}%
  \let\thefootnote\svthefootnote%
}

\DeclareMathOperator*{\argmax}{arg\,max}

\AtBeginDocument{%
  \providecommand\BibTeX{{%
    \normalfont B\kern-0.5em{\scshape i\kern-0.25em b}\kern-0.8em\TeX}}}

\setcopyright{acmcopyright}
\copyrightyear{2022}
\acmYear{2022}
\setcopyright{acmcopyright}
\setcopyright{acmcopyright}\acmConference[MM '22]{Proceedings of the 30th ACM International Conference on Multimedia}{October 10--14, 2022}
{Lisboa, Portugal}
\acmBooktitle{Proceedings of the 30th ACM International Conference on Multimedia (MM '22), October 10--14, 2022, Lisboa, Portugal} \acmPrice{15.00}
\acmDOI{10.1145/3503161.3548289} \acmISBN{978-1-4503-9203-7/22/10}
\begin{document}

\title{Angular Gap: Reducing the Uncertainty of Image Difficulty through Model Calibration}

\author{Bohua Peng}
\affiliation{%
  \institution{Imperial College London}
  \city{London}
  \country{UK}
}
\email{bohua.peng19@imperial.ac.uk}

\author{Mobarakol Islam\textsuperscript{\Letter}}
\affiliation{%
  \institution{Imperial College London}
  \city{London}
    \country{UK}
  }
\email{m.islam20@imperial.ac.uk}

\author{Mei Tu}
\affiliation{%
  \institution{Samsung Research}
    \city{Beijing}
    \country{China}
  }
\email{tumei@outlook.com}

\renewcommand{\shortauthors}{Bohua Peng, Mobarakol Islam, \& Mei Tu.}

\begin{abstract}
   Curriculum learning needs example difficulty to proceed from easy to hard. However, the credibility of image difficulty is rarely investigated, which can seriously affect the effectiveness of curricula. In this work, we propose Angular Gap, a measure of difficulty based on the difference in angular distance between feature embeddings and class-weight embeddings built by hyperspherical learning. To ascertain difficulty estimation, we introduce class-wise model calibration, as a post-training technique, to the learnt hyperbolic space. This bridges the gap between probabilistic model calibration and angular distance estimation of hyperspherical learning. We show the superiority of our calibrated Angular Gap over recent difficulty metrics on CIFAR10-H and ImageNetV2. We further propose Angular Gap based curriculum learning for unsupervised domain adaptation that can translate from learning easy samples to mining hard samples. 
   We combine this curriculum with a state-of-the-art self-training method, Cycle Self Training (CST). 
  The proposed Curricular CST learns robust
  representations and outperforms recent baselines on Office31 and VisDA 2017. 
\end{abstract}

\maketitle
\begin{figure}[t]
\begin{center}
\includegraphics[width=\linewidth]{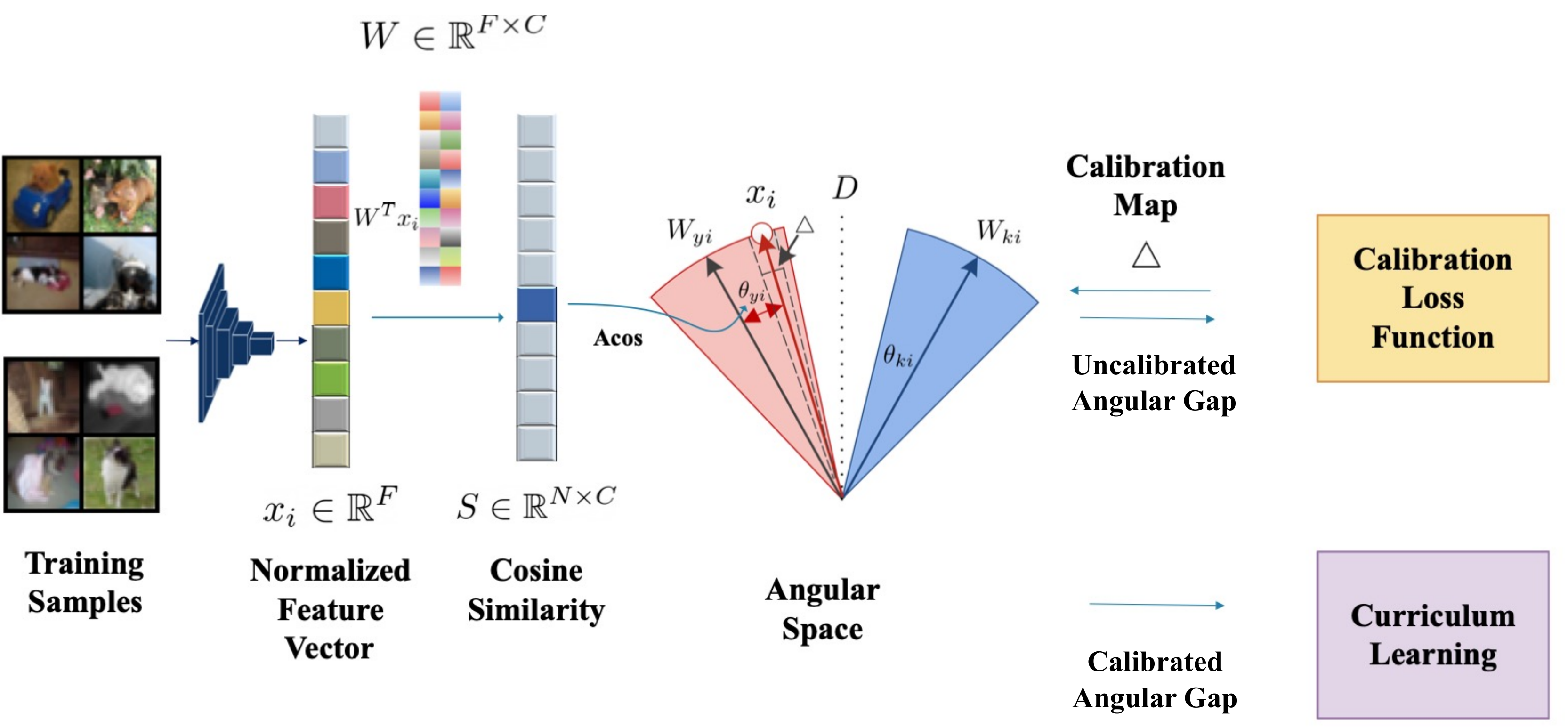}
\end{center}
\caption{
An overview of our Angular Gap image difficulty quantification framework. In the training stage, deep neural networks learn image vectors and class vectors in an angular space with label information, and output raw difficulty scores based on angles. Difficulty scores are then calibrated on a hold-out validation set. In the test stage, the proposed framework can output example difficulty for downstream tasks.
}
\label{fig:teaser}
\end{figure}
\section{Introduction}
Ascertaining example difficulty is a critical problem to curriculum learning and self-paced learning, in that curricula rank training samples by difficulty and proceed from easy to hard. In the context of image classification, a natural idea is to quantify such difficulty with human selection frequency\cite{recht2019imagenet}, i.e., the fraction of annotators selecting a sample for its target class. However, human labelling effort is not scalable to get fine-grained image difficulty. To measure the difficulty of 10,000 CIFAR10 images, CIFAR10-H\cite{Battleday2020CapturingHC} recruits 2570 annotators to perform 511,400 trials, with an average of 51 human decisions per image, not including a considerable amount of practice and attention checks. Hence, automating difficulty estimation is crucial to applying curriculum learning to large scale datasets. \\
Probabilistic models are particularly compelling for this automatic estimation demand because of their consistency towards noisy image contents and uncertain labels regularly presented in large scale datasets. Early works have characterized image difficulty with maximum confidence or classification margin, the difference between the predicting probability of the correct class and the largest among others. However, difficulty measurers based on modern neural networks have a reputation of being poorly calibrated. Extensive research have shown that the negative log-likelihood can easily overfit training samples, pushing average predicting probability away from accuracy\cite{guo2017calibration, kull2019beyond}. This suggests considerable uncertainty of softmax probabilities, and imprecise difficulty measurement undermines the performance of curriculum learning. While probability estimation deteriorates, final classification results actually improve\cite{guo2017calibration}. Very recently, deep ensemble methods \cite{jiang2020characterizing, baldock2021deep} measure example difficulty with agreement either from last layers' predictions or from intermediate layers' predictions. Reducing estimation uncertainty with ensembling requires selected treatments and controls such as  architectures, number of submodels, and number of data splits. In this work, we show faithful image difficulty can be efficiently estimated by deep metric learning.\\
Hyperspherical learning\cite{liu2017deep}, a weakly supervised learning framework, groups instances of the same concept together and pushes instances of different concepts apart by enforcing angular discrimination during training.
 The framework allows for more robust similarity estimation and has improved representation learning in both computer vision\cite{deng2019arcface} and natural language understanding\cite{feng2020language}. Specifically, samples and classes are projected as vectors with constant norms in a hyperbolic space. The normalization operation creates {\itshape"radial"} feature
distributions, and the corresponding cosine similarity has been proved to be robust for many downstream tasks\cite{chen2020simple}. The benevolent properties of hyperspherical similarity estimation give us motivation for difficulty estimation. Angular visual hardness (AVH)\cite{chen2020angular} initially defines difficulty as the angular distance to its label class normalized by the sum of angular distances to all classes. However, a limitation of this difficulty is that significant angular information can be flushed away by the accumulation of imprecise angular distances. For instance, if an image shows a tabby cat, the distance to its class vector is washed out by distances to unrelated classes, e.g., goldfish or sailboat, resulting in example difficulty with high variance. Based on the assumption that more probable predictions are better calibrated\cite{kull2019beyond}, we propose a new difficulty defined as the difference between angular distances of the label class and the smallest of other classes as illustrated in Figure~\ref{fig:teaser}.
Additionally, we introduce a multi-level calibration method to reduce estimation uncertainty through post-training calibration.\\
In summary, our contributions and findings are summarized below:
1. )We propose Angular Gap to measure example difficulty for designing a curriculum learning scheme.
2. ) We develop multilevel calibration techniques with global and class-wise calibration to produce accurate uncertainty for Angular Gap.
3.) We propose a smooth transfer learning curriculum and integrate CST with calibrated Angular Gap for the unsupervised domain adaptation task
4. )We extensively validate calibrated Angular Gap on several SOTA methods and datasets of unsupervised domain adaption and the results suggest the superior performance.
\section{Related works}
\freefootnote{Code available at \url{https://github.com/pengbohua/AngularGap}.}
\label{sec:related}
\paragraph{Image difficulty}A wide range of researchs show images possess different amounts of difficulty. It takes tremendous efforts to quantify human perceptual image difficulty. 
Recently, a line of works model difficulty with
the "learning dynamics" of labelling functions.
Forgetting events\cite{Toneva2019AnES} relate example difficulty to catastrophic forgetting \cite{french1999catastrophic} by measuring the occurrence of a sample being forgotten during training. The measurement is generalized from discrete domains to continuous domain by averaging the results of ensembles. C-score \cite{jiang2020characterizing} designs a Monte Carlo method to estimate difficulty w.r.t the probability of correct generalization. Prediction depth \cite{baldock2021deep} employs an ensemble of k-NN classifiers to output intermediate predictions, and defines difficulty as the earliest layer where subsequent intermediate predictions converge. However, difficulty measured by deep ensembles rely on selected treatments and controls such as architectures, data splits and ensembling strategies. Recently, Angular Visual Hardness \cite{chen2020angular} initially tries to model image difficulty with angular distance predicted by a single neural network.
In this work, we reinforce this idea with hyperspherical learning\cite{liu2017deep} that emphasizes angular discrimination and ascertain difficulty with model calibration.\\
\paragraph{Uncertainty estimation}
The shared goal of uncertainty estimation and model calibration is to provide trustworthy model confidence for decision making. Expected calibration error (ECE) and reliability diagrams are standard metrics to measure model calibration\cite{platt1999probabilistic}. Recently, deep ensembles\cite{lakshminarayanan2017simple, jiang2020characterizing} become popular methods for visual uncertainty estimation due to less correlation between individual models. The major drawback is their heavy computational overheads. To ascertain the model confidence of a single model, another line of research focuses on post-training calibration. Platt scaling\cite{platt1999probabilistic} is a test-by-time parametric approach that rescales output logits with an extra linear layer trained on a hold-out validation set. Temperature Scaling (TS)\cite{guo2017calibration} simplifies this approach with a single learnable parameter. Most recently, variants of Platt scaling\cite{ kull2019beyond} and class-wise TS \cite{islam2021class} present better calibration performance over vanilla TS. However, Dirichlet Calibration\cite{kull2019beyond} claims that Temperature Scaling mainly focuses on the maximum probability instead of predictions of all classes, which aligns with \cite{nixon2019measuring}. In this work, we opt to revisit model calibration and predict plausible similarity in a hyperbolic space.\\
\paragraph{Curriculum learning}
Curriculum learning\cite{Bengio2009CurriculumL} is a paradigm that favors learning along a curriculum of examples from easy to hard. Starting from this general idea, self-paced learning \cite{kumar2010self} implements an automatic curriculum that considers examples with small loss as representative examples. With recurrent neural network,  MentorNet\cite{jiang2018mentornet} combines the best of curriculum learning and self-paced learning with a teacher-student architecture that supervises the training of base networks by learning a data-driven curriculum. 
In the context of supervised learning, deep neural networks learn transferrable features from representative examples before overfitting specific features\cite{jiang2020characterizing}. We extend the above ideas to hyperspherical learning and propose curricula that prioritize large Angular Gap.\\
\paragraph{Unsupervised domain adaptation}
Unsupervised domain adaptation (UDA) presents a challenging transfer learning problem where data from the source domain are labeled while data from the target domain are unlabeled. A shared assumption between feature alignment methods\cite{ganin2015unsupervised, zhu2020deep}and self-training algorithms\cite{zou2018unsupervised, na2021fixbi} is that shared knowledge exists between domains, which allows for the same labelling function. On the one hand, shared knowledge exists as similar features between domains in the feature adaptation literature\cite{ghifary2014domain, long2015learning}. On the other hand, shared knowledge is modelled by the parameters of feature extractors in the self-training algorithms\cite{zou2018unsupervised, zou2019confidence,liu2021cycle}. Using source models to label target data, CBST\cite{zou2018unsupervised} initially performs pseudo-label selection with class-wise confidence thresholds, which is then improved by confidence regularization as CRST\cite{zou2019confidence}. To handle large domain discrepancy, FixMatch\cite{sohn2020fixmatch} applies a pair of weak and strong data augmentations to target image and enforce consistency regularization when the weakly-augmented image prediction is confident. FixBi\cite{na2021fixbi} uses a fixed mixup ratio to train twin feature extractors as {\itshape"bridges"} between domains.
As an alternative, curriculum learning has been applied to domain adaptation from task-level\cite{zhang2017curriculum} or instance-level\cite{shu2019transferable} using feature adaptation. We borrow these ideas and propose a transfer learning curriculum to create infinite bridges that gradually decrease discrepancy between different domains. Different from existing works, our transfer learning curriculum does not purely focus on easy samples, but choose the optimal from searching.
\begin{figure}
    \centering
    \includegraphics[width=0.9\linewidth]{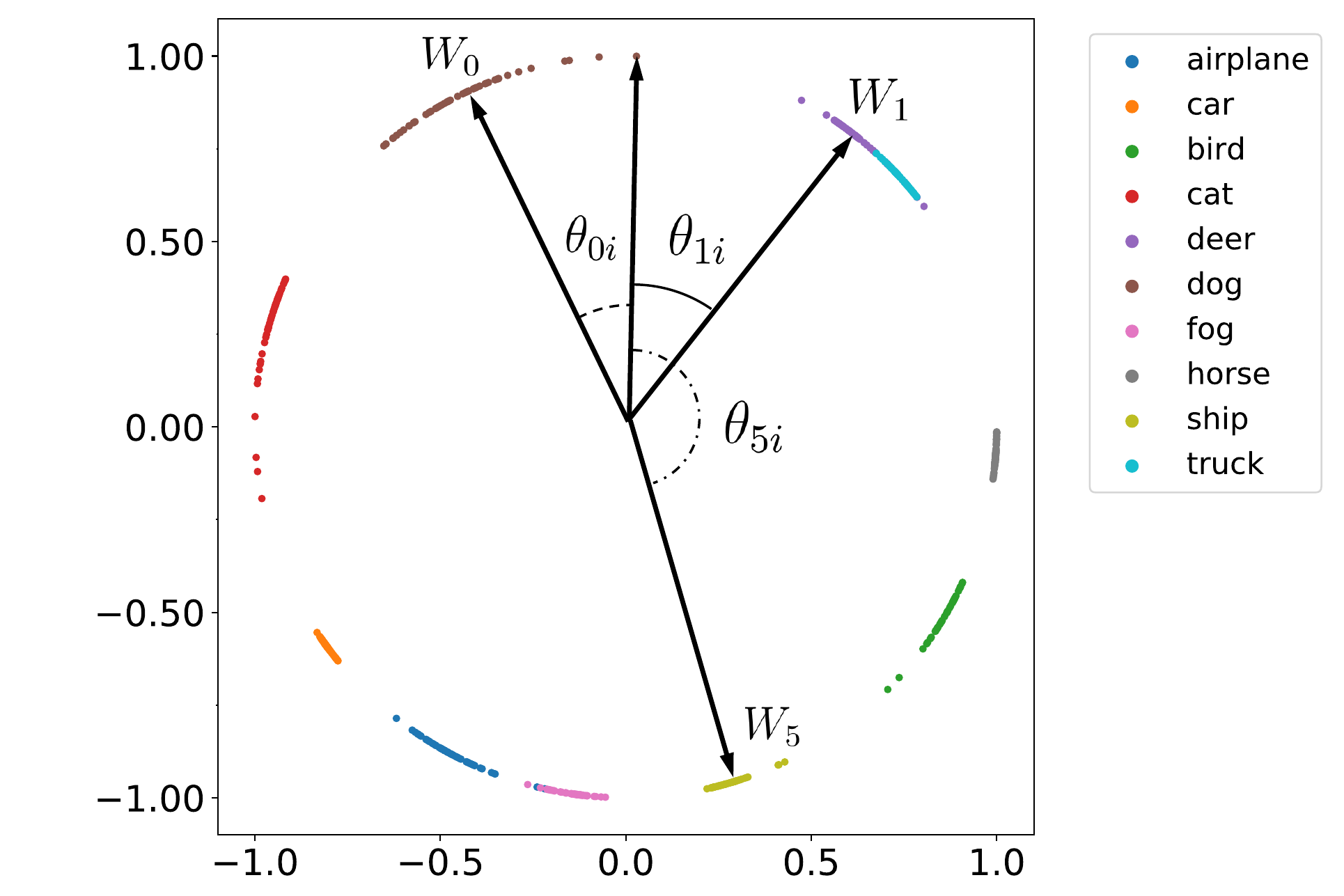}
    \caption{An example of Angular Visual Hardness (AVH). Three out of ten angles are shown for visual clarification.}
    \label{fig:avh}
\end{figure}
\section{Preliminary}
\paragraph{Angular Visual Hardness}Figure \ref{fig:avh} shows an example of image difficulty measured by Angular Visual Hardness (AVH). This metric is automatically estimated by a neural network formed by a feature extractor and a linear classifier. For an individual image, AVH is defined as the angular distance between its feature vector and label weight vector divided by the sum of angular distance between its feature vector and all class weights.
\begin{gather}
    \textit{AVH}(\mathbf x_i)=\frac{\mathcal{A}(\mathbf{x}_i, \mathbf{w}_{yi})}{\sum_{k=1}^{C}\mathcal{A}(\mathbf{x}, \mathbf{w}_k)}\\
    \mathcal{A}(\mathbf {x}_i, \mathbf{w}_k) = \arccos{(\frac{\mathbf{x}_i^T \mathbf{w}_k}{\left\|{\mathbf x_i}\right\|\left\|{\mathbf{w}_k}\right\|})}
\end{gather}
where $\mathbf x_i\in\mathbb{R}^{d}$ denotes the $d$ dimensional image feature extracted by a backbone. The image is categorized into one of $C$ classes and labeled as $y$. $\mathbf{w}_k\in\mathbb{R}^{d}$ is the $k-$th column of the linear classifier's weight $\mathbf W$. \\
\paragraph{Subdomain feature alignment} Deep Subdomain Adaptation Network\cite{zhu2020deep} performs fine-grained feature alignment by dynamically weighing up samples from less representative classes. In their method, local maximum mean discrepancy (MMD) can be measured as follows
\begin{gather}
    d_{\mathcal{H}}(P, Q)\triangleq\sum_{k=1}^{C}u_k\left\|\frac{1}{|X^S|}\sum_{\mathbf{x}_s\in \mathcal{S}}\phi(\mathbf{x}_s)-\frac{1}{|X^T|}\sum_{\mathbf{x}_t\in \mathcal{T}}\phi(\mathbf{x}_t)\right\|_{\mathcal{H}}
\label{eq:dsan}
\end{gather}
where $u_k$ is the class ratio to characterize subdomains defined in \cite{zhu2020deep}. $x^s$ and $y^s$ are features and labels from the source domain $\mathcal{S}$, and $x^t$ are unlabeled features from the target domain $\mathcal{T}$. $\phi$ are kernel functions that measure the distance between source feature $\mathbf{x}_s$ and target feature $\mathbf{x}_t$ on a Hilbert space. Deep features are optimized with the classification loss and this transfer loss.\\
\paragraph{Cycle Self-Training}
As a state-of-the-art single perspective UDA method, Cycle Self-training (CST)\cite{liu2021cycle} contains an inner loop and an outer loop. Both loops share the same feature extractor but have their own classifier. With input augmentations and consistency regularization, the inner loop focuses on correctly predicting target samples. The outer loop updates feature representations to reduce the difference between source classifier and target classifier.
\begin{figure*}
\begin{center}
\includegraphics[width=0.9\linewidth]{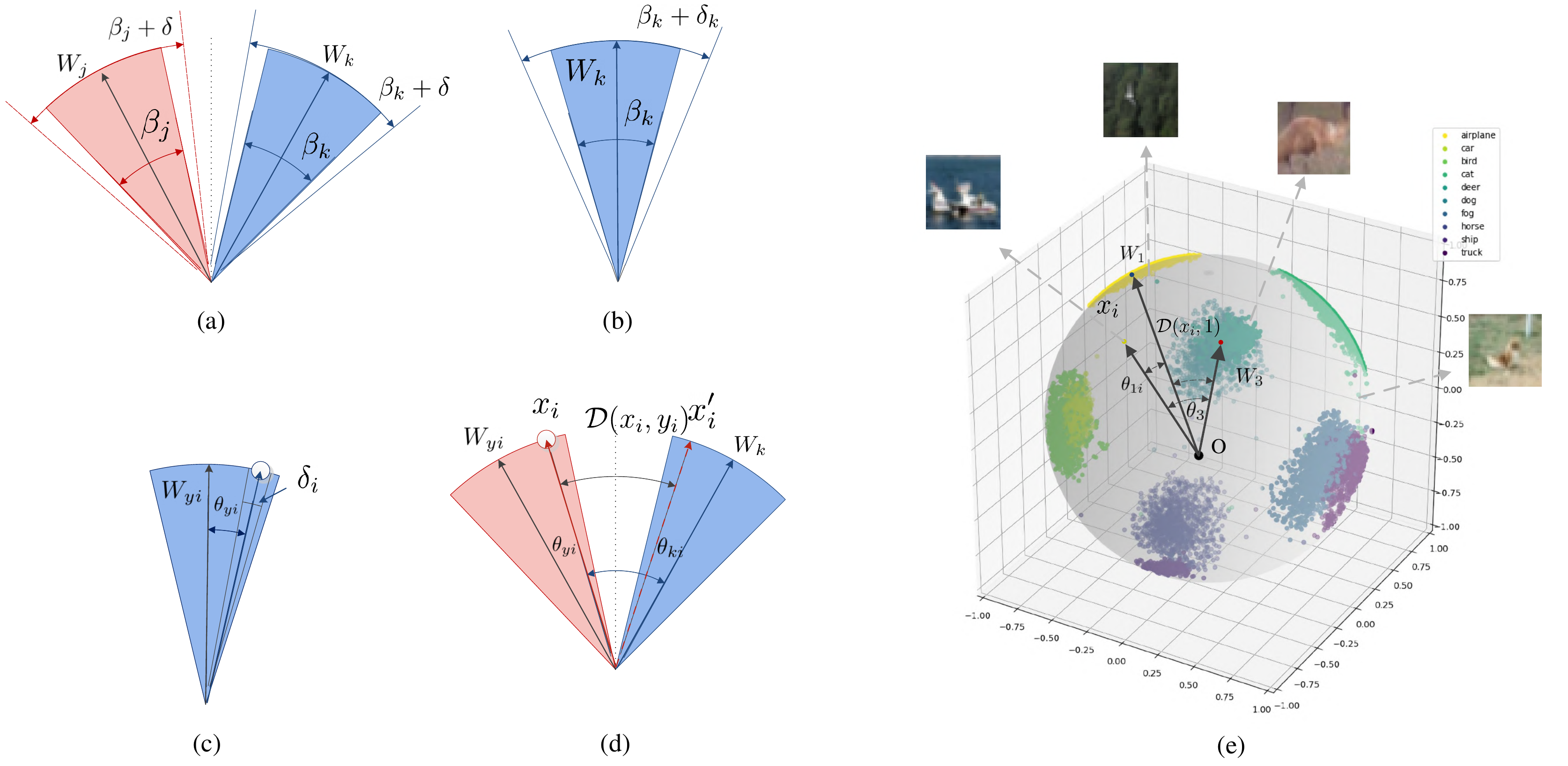}
\end{center}
\caption{
Geometric interpretation of model calibrations on a hyperbolic space. A neural network learns class weights $W$, features $x$ and angles $\beta$ and $\theta$ during training. These angles are rectified by $\delta$ with post-training calibration. (a) Global calibration adds a small angle $\delta$ to all samples simultaneously, hence increasing the angular range of class j from $\beta_j$ to $\beta_j+\delta$. (b) Classwise calibration learns a vector $\boldsymbol{s}_d$ that adds $\delta_k$ to the angles of class k. (c) Model calibration rectifies an individual angle $\theta_yi$ by $\delta_i$. (d) An angular gap is the difference between angle $\theta_{y_i}$ of the label class and the smallest angle $\theta_k$ among other classes. (e) We visualize the image difficulty of CIFAR10 measured by Angular Gap $\mathcal{D}(x_i, y_i)$ on a 3D globe.
}
\label{fig:overview}
\end{figure*}
\section{Calibrated Angular Gap}
In the context of image classification, we propose Calibrated Angular Gap to estimate example difficulty for curriculum learning. The learnt difficulty metric is based on the angular distance between the feature vectors and the class-weight vectors predicted by hyperspherical learning. In the standard curriculum learning\cite{Bengio2009CurriculumL}, we train the model with easier samples determined by Angular Gap, and then gradually feed harder samples. For domain adaptation, we propose a novel curriculum to work with Angular Gap, which provides a smooth transition between adapting easy samples and hard sample mining. We combine this method with cycle self-training (CST).
\subsection{Angular Gap}\label{sec:ag}
Example difficulty can be considered as modelling {\itshape "similarity"} between examples and abstract concepts. The abstract concepts can be class labels, prototypes, or even text descriptions. For simplicity, we define a new difficulty, Angular Gap, measured as the difference between the similarity to its label class and the largest similarity of all classes. This definition is based on the assumption that larger cosine similarities are more precisely estimated than smaller ones. For example, when searching with an image of a tabby cat, one can probably get many of its kind and some tiger cats because of their common visual properties. \\
\textbf{Definition1} Formally, we represent Angular Gap as
\begin{gather}\label{eq:definition}
    \mathcal{D}(\mathbf x, \mathbf y)= sim(\mathbf x, \mathbf{w}_y) - \argmax_{k\neq y}sim(\mathbf {x}, \mathbf{w}_k) \\
    sim(\mathbf x, \mathbf{w}_k) =\cos\theta_{k}= \frac{\mathbf{x}_i^T \mathbf{w}_k}{\left\|{\mathbf x}\right\|\left\|{\mathbf{w}_k}\right\|},
\end{gather}
where $\theta_{k}$ is the angle between $\mathbf x$ and $\mathbf{w}_k$.\\
Following common practice of image recognition\cite{wang2018cosface, deng2019arcface}, we emphasize angular discrimination on the hyperbolic space with normalized softmax loss (NSL) and feature norm rescaling represented as
\begin{equation}
L_{NSL} = - \frac{1}{N}\sum_{i=1}^N\log \frac{\exp(s\cdot \cos\theta_{y_i}
)}{\sum_{k=1}^{C} \exp(s\cdot \cos\theta_{k})}
\end{equation}
$s$ denotes the scaling factor that rescales feature norms to a constant. Unlike \cite{deng2019arcface} that inserts a geodesic margin between the sample and its class center, here we remove the margin to achieve better generalisation. Note that feature normalization has projected features to a hypersphere with a radius of $s$.
\subsection{Multilevel model calibration}
In our empirical study, feature norms increase continuously yet slowly when training with NSL, indicating negative log-likelihood overfitting training data. This is partially because the cosine similarity get minimized when feature norms increase. Although overfitting may increase test accuracy, uncertain similarities harm difficulty estimation. As shown in Figure ~\ref{fig:overview}, we handle this problem with model calibration from a global level, a class-wise level and an instance level. In general, our idea is to learn multiplicative calibration functions that refine the angles on a hyperbolic space.
\begin{align}
    \cos(\boldsymbol{\theta} +\boldsymbol{\Delta})&=\cos\boldsymbol{\Delta}\cos\boldsymbol{\theta}-\sin\boldsymbol{\Delta}\sin\boldsymbol{\theta}\\
    &\approx \cos\boldsymbol{\Delta}\cos\boldsymbol{\theta}-\boldsymbol{\Delta}\sin\boldsymbol{\theta}\\
    &= \varphi(\boldsymbol{x},\boldsymbol{\theta})\cos\boldsymbol{\theta}
\end{align}
With small-angle approximation, the nonlinear calibration function $\varphi(\boldsymbol{x},\boldsymbol{\theta})$ adds or remove a small angle $\delta$ from the original prediction. \\
\textbf{Global calibration} Global calibration expands or shrinks all angles on the hyperbolic space simultaneously with a single learnable parameter $s_t$. This requires a validation dataset with samples $x_j\in\chi$ and labels $y_j\in\mathcal{Y}=\left \{,...C\right \}$. The loss function for this calibration method is 
\begin{gather}
\min_{s_t}L_{global} = - \frac{1}{N}\sum_{i=1}^N\log \frac{\exp( s\cdot(s_t \cdot \cos\theta_{y_i}
))}{ \sum_{k=1}^{C} \exp (s\cdot(s_t \cdot \cos\theta_{k}))},\\
\cos\xi _k=s_t \cdot \cos\theta_{k},
\label{eq:globalloss}
\end{gather}
$s_t$ is an additional parameter learnt during post-training calibration. $\cos\xi _kg$ is the refined angular distance at global level. \\
\textbf{Class-wise calibration} A single parameter is not enough to give us precise example difficulty calibration. To capture class-level difficulty exhibited in behavioral datasets, we let the neural network learn a vector $\mathbf s\in\mathbb{R}^{C}$ that equally rescales angles with another calibration loss function defined as 
\begin{gather}
L_{class} = - \frac{1}{N}\sum_{i=1}^N\log \frac{\exp(s\cdot(s_{y_i} \cdot \cos\theta_{y_i}
))}{ \sum_{k=1}^{C} \exp(s\cdot(s_k \cdot \cos\theta_{k}))},\\
\cos\xi _k=s_k \cdot \cos\theta_{k},
\label{eq:classloss}
\end{gather}
where $s_k$ is the $k$th entry of the vector $\mathbf s$ corresponds to class k. $s_{y_i}$ rescales the angular distance to the label class. $\cos\xi _k$ is the calibrated angular distance at class level.\\
\begin{equation}
    \mathcal{D}^*(\mathbf x, \mathbf y)= \cos \xi_y - \argmax_{k\neq y}\cos \xi_k
\end{equation}
 Augmented image difficulty $\mathcal{D}^*$ can be computed by replacing the directly measured similarities $\cos\theta_k$ with refined similarities $\cos\xi_k$, as done with global calibration and class-wise calibration. The proposed calibration methods are natural extensions of Temperature Scaling and Vector Scaling\cite{guo2017calibration} in the hyperspherical setting.
\begin{table*}[]
    \centering
    \caption{Accuracy (\%) of standard curriculum learning guided by example difficulties on CIFAR10-H. Our methods outperform AVH by 1\% and are on par with C-score. Note that C-score was built with a deep ensemble with selected
    data splits, while we train a single AngularGap model from scratch. Correlations with human selection frequency are measured with Spearman's rank and Kendall's Tau, with $p<0.001$ for all experiments. Calibration is empirically measured with ECE(\%). Global calibration, Class-wise calibration and Temperature Scaling calibration is represented by ${\rm Global}$, ${\rm Class}$-$\rm{wise}$ and ${\rm TS}$ respectively.}\label{tab:main}.
    \begin{tabular}{lccccccc}
    \toprule
      Methods  & Spearman's rank& Kendall's Tau&ECE&Top-5 acc.&Top-1 acc.\\
      \midrule
      Maximum Confidence& 0.266$\pm$0.006        & 0.148$\pm0.004$&11.3$\pm$0.2&94.5$\pm$0.3&74.8$\pm$2.1 \\
      Maximum Confidence$\rm_{TS}$&0.273$\pm$0.004        & 0.145$\pm0.004$&9.1$\pm$0.2&94.5$\pm$0.3&75.3$\pm$2.1 \\
      Classfication Margin\cite{Toneva2019AnES}& 0.279$\pm$0.006   & 0.142$\pm$0.004&11.3$\pm$0.2&94.6$\pm0.3$&75.3$\pm$2.0  \\
      Classfication Margin$\rm_{TS}$& 0.283$\pm$0.006   & 0.242$\pm$0.004&9.0$\pm$0.2&94.9$\pm0.3$&75.7$\pm$1.3 \\
      MC-Dropout\cite{gal2016dropout}&0.256$\pm$0.007&0.176$\pm$0.006&9.4$\pm$0.4&95.7$\pm$0.3&77.0$\pm$0.5\\
      AVH\cite{chen2020angular} & 0.368$\pm$0.006      & 0.258$\pm$0.004&8.2$\pm$0.2&98.2$\pm$0.3&81.2$\pm$1.0\\
       AVH$_{\rm Global}$& 0.376$\pm$0.003      & 0.263$\pm$0.003&\underline{7.5$\pm$0.2}&98.6$\pm$0.2&81.3$\pm$0.7\\
       AVH$_{\rm Class-wise}$& 0.377$\pm$0.003      & 0.265$\pm$0.002&\textbf{7.4$\pm$0.2}&98.6$\pm$0.2&81.4$\pm$0.6\\
      Forgetting Events\cite{Toneva2019AnES}& 0.260$\pm0.003$ & 0.187$\pm0.002$&11.5$\pm$0.5&98.0$\pm$0.4&78.9$\pm$1.2\\
      C-score\cite{jiang2020characterizing} & 0.316$\pm0.001$& 0.243$\pm0.001$&9.8$\pm$0.3&\textbf{99.0$\pm$0.1}&\textbf{82.4$\pm$0.4}\\ 
Prediction Depth\cite{baldock2021deep} &0.290$\pm0.001$ & 0.183$\pm0.001$&9.8$\pm$0.3&98.5$\pm$0.2&81.2$\pm$0.4\\
\midrule
Angular Gap (Ours) & 0.378$\pm$0.003 &0.265$\pm$0.003&8.2$\pm$0.2&98.6$\pm$0.2&82.0$\pm$0.6\\
Angular Gap$_{\rm Global}$& \underline{0.382$\pm$0.003} &\underline{0.268$\pm$0.002}&\underline{7.5$\pm$0.2}&98.8$\pm$0.2&82.3$\pm$0.4\\
Angular Gap$_{\rm Class-wise}$ & \textbf{0.384$\pm$0.002} &\textbf{0.269$\pm$0.002}&\textbf{7.4$\pm$0.2}&\underline{98.9$\pm$0.2} &\textbf{82.4$\pm$0.4}\\
    \bottomrule
    \end{tabular}
\end{table*}
 \subsection{Curricular Cycle Self-Training}
\begin{figure*}[ht]
\centering
\includegraphics[width=0.75\linewidth]{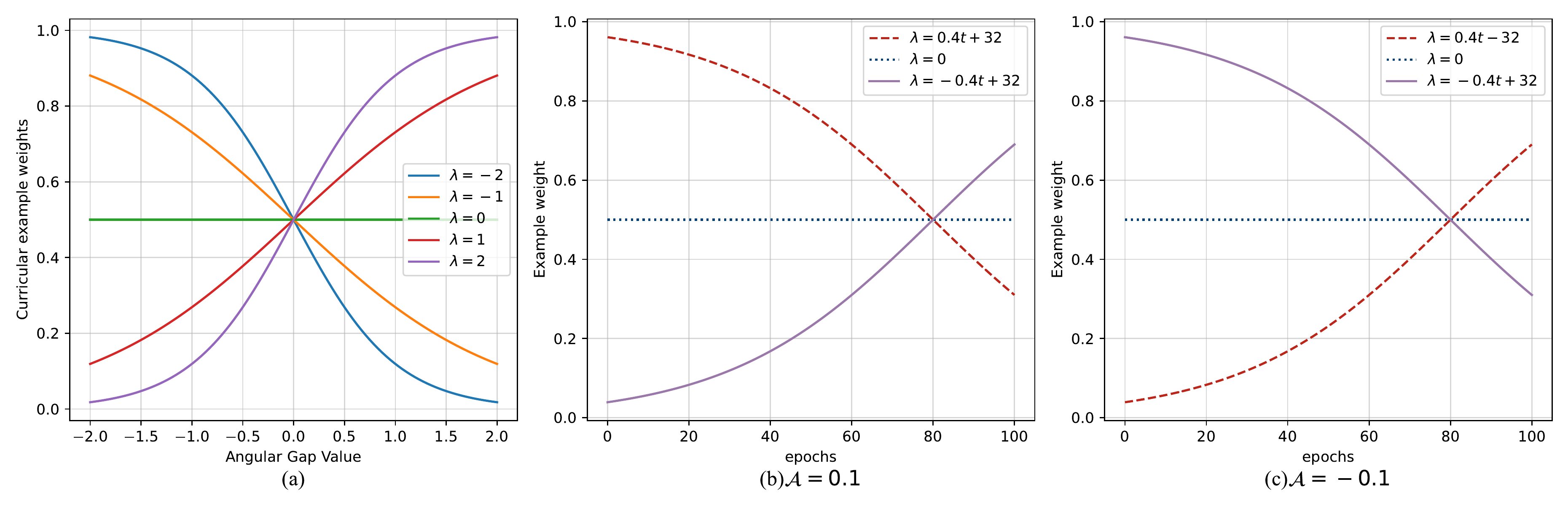}
\includegraphics[width=0.75\linewidth]{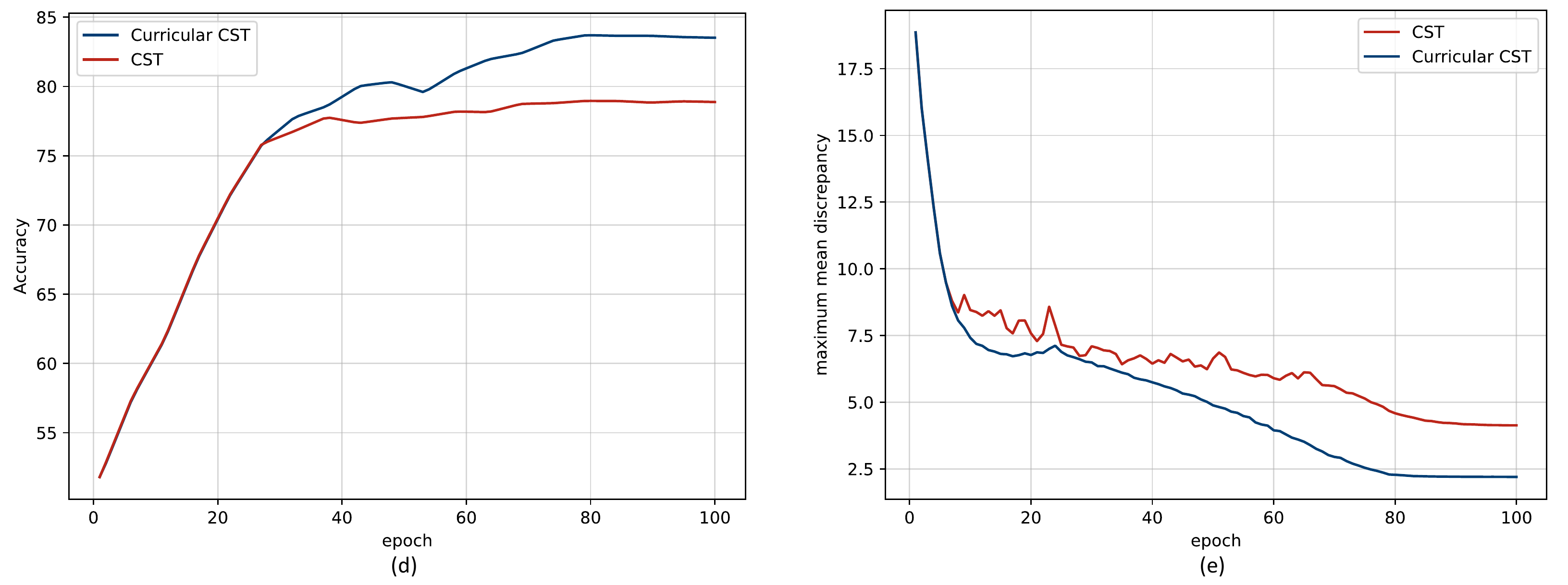}
\caption{Curricular cycle self-training. The pacing function $\lambda$ schedules example weights as training proceeds. The optimal transfer learning curriculum, shown as the purple line, decreases the importance of an easy sample (b) and increases the importance of a hard sample (c) as training proceeds. Note that the purple curriculum enforces easy-to-hard at the beginning and hard sample mining at the end. (d) and (e) compare model accuracy and the maximum mean discrepancy between CST (red) and Curricular CST(blue) for all target data points.}
\label{fig:sym}
\end{figure*}

 With Angular Gap, we facilitate domain adpation by defining a novel curriculum that prioritizes alignment on easy samples that contain general necessary knowledge, and gradually focus on hard samples that contain specific knowledge. Because alignment can be measured by fixed kernel functions or neural classifiers, we combine Angular Gap with DSAN\cite{zhu2020deep} and CST\cite{liu2021cycle}, and propose Curricular DSAN and Curricular CST respectively.\\
In Figure~\ref{fig:sym}, we design a novel curriculum that control example weights according to pacing functions and example difficulty with sigmoid functions. We choose sigmoid functions to work with Angular Gap $\mathcal{A}(\boldsymbol{W},x_s, y_s))$ because this combination allows for efficiently searching pacing functions $\lambda$ in a symmetric space. Moreover, this curriculum can smoothly transfer between easy-to-hard and hard sample mining.
\begin{gather}
    d_s = \sigma(\lambda\cdot\mathcal{D}(x_s, y_s)),\\
\lambda_{(a,b)}(t) = N\frac{1-b}{aT}t + Nb.
\label{eq:cl}
\end{gather}
where $d_s$ is example weight ranged between 0 and 1. $a$ and $b$ denote the parameters of pacing functions $\lambda$. $t$ is the current time step and $T$ is the number of total iterations.\\
\paragraph{Curricular DSAN}We want to learn generalizable features by prioritizing alignment of easy samples with the same class.  We augment the feature alignment process with a dynamic discrepancy scoring function defined as
\begin{gather}
    d_{\mathcal{H}}(P, Q)\triangleq\sum_{k=1}^{C}u_k\left\|\frac{d_s}{|X^S|}\sum_{x_s\in S}\phi(x_s)-\frac{1}{|X^T|}\sum_{x_t\in T}\phi(x_t)\right\|_{\mathcal{H}}.
\end{gather}
\paragraph{Curricular CST} We apply Angular Gap to CST to improve the optimization of the outer loop. Our curriculum prioritizes model updates for transferring and scoring the features of easy samples, yielding more robust representations for pseudo-labels generation of the inner loop. To this end, we add example weights to the reverse step as follows,
\begin{equation}
L_{rev}\triangleq\sum_{s\in S}\sum_{t\in T}d_s\left\|sim(x_s,x_t)\hat{y}_t-y_s\right\|^2.
\end{equation}
This term reduces cross-domain discrepancy by explicitly transforming pseudo labels $\hat{y}$ to the source domain, and updates deep representations learnt by the mutual feature extractor.
\section{Experiments and Results}
We have designed the experiments to evaluate our proposed methods by evaluating the following hypotheses:
\begin{itemize}
    \item More credible example difficulty: We evaluate the credibility of learnt Angular Gap by analyzing their correlations with human selection frequency on CIFAR10-H and ImageNetV2.
    \item Better curriculum learning: We evaluate the performance of networks when guided with Angular Gap under the standard curriculum learning framework \cite{Bengio2009CurriculumL}.
    \item More robust representations: We evaluate how the network is able to learn better generalizable representations with the proposed curriculum on the task of unsupervised domain adaptation. 
\end{itemize}
\subsection{Datasets}
CIFAR10\cite{krizhevsky2009learning} and ISLRVC 2012 (ImageNet) \cite{krizhevsky2012imagenet} are standard benchmarks for image classification . For human evaluation, CIFAR10-H\cite{Battleday2020CapturingHC} and ImageNetV2\cite{recht2019imagenet} are two recently popular behavioral datasets that report human selection frequency. Human selection frequency models instance-level difficulty with the faction of people that correctly classify an image. CIFAR10-H is composed of 10,000 images from 10 classes with 511,400 decisions given by 2570 annotators. ImageNetV2 is composed of 10,000 large-scale images from 1000 classes, where each image is labelled by more than 10 annotators.\\
For domain adaptation, we consider  Office-31\cite{saenko2010adapting} and VisDA 2017\cite{peng2017visda} as standard benchmarks. Office-31 consists of images of 31 classes from three domains - Amazon (A), DSLR (D) and Webcam (W). Each domain has 2, 817, 498 and 795 images respectively. We compare our Curricular DSAN and Curricular CST with recent baselines across all six transfer learning tasks. VisDA 2017 considers 152, 409 labeled synthetic images as the source domain and 55, 400 unlabeled real-world images as the target domain.
\begin{table}[]
\centering
\caption{
\small{Correlations between difficulty and HSF on ImageNetV2.}
} 
\label{tab:imagenetv2}
\resizebox{\linewidth}{!}{ 
\begin{tabular}{@{}lcccc@{}}
\toprule
        & \multicolumn{2}{c}{Spearman's rank}& \multicolumn{2}{c}{Kendall's Tau} \\
\cmidrule{2-5}
&  $\rho$ & p-value & $\tau$ & p-value \\
\midrule
Maximum Confidence & 0.273 &$<$0.001  & 0.201 &$<$0.001 \\
Classification margin & 0.275&$<$0.001   & 0.204 &$<$0.001 \\
Forgetting Events\cite{Toneva2019AnES} & 0.260  &0.048 & 0.187 &0.054 \\
Prediction Depth\cite{baldock2021deep}& 0.308  &$<$0.001 & 0.192 &$<$0.001\\
Classification margin (TS) & 0.293&$<$0.001   & 0.242 &$<$0.001 \\
AVH (Global)\cite{chen2020angular} & 0.377&$<$0.001      & 0.257 &$<$0.001\\
\midrule
Angular Gap (Global) & \underline{0.379}&$<$0.001 &\underline{0.269}&$<$0.001\\
Angular Gap (Class) & \textbf{0.382}&$<$0.001 &\textbf{0.271}&$<$0.001\\
\bottomrule
\end{tabular}
} 
\end{table}
\subsection{Implementation details}\label{sec:imp}
\paragraph{Image difficulty estimation}To measure image difficulty, we employ popular convolutional neural networks. We emphasize Angular discrimination with normalized softmax loss and a rescaling factor $s$ of 30. For data prepossessing, we follow PyTorch examples\cite{NEURIPS2019_9015} to generate random image crops. We train the models from scratch on CIFAR datasets for 100 epochs with a batch size of 128. We set the initial learning rate as 0.1 and a cosine learning rate annealing strategy. On ImageNet, we finetune pretrained ResNet50 with SGD optimization and set hyperparameters as stated in PyTorch examples\cite{NEURIPS2019_9015}. \\
\begin{figure}[htbp]
\centering
\includegraphics[width=\linewidth]{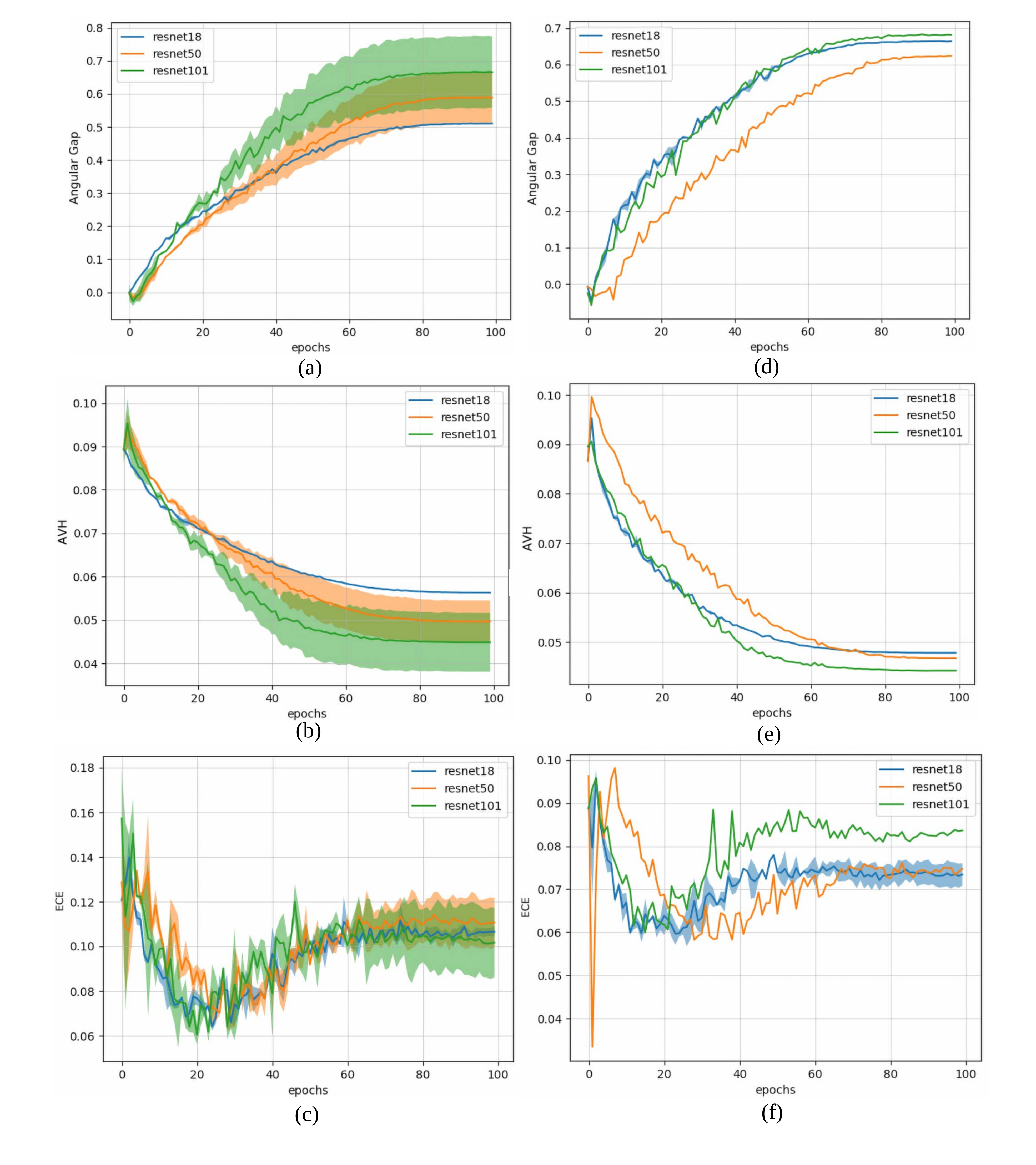}
\caption{Comparison between Angular Gap and AVH before and after class-wise calibration on CIFAR10-H. Shadows in the plots denote the corresponding standard deviations. The first column shows the training dynamics of uncalibrated Angular Gap (a) and AVH (b) and calibration (c) of the hyperspherical model. The second column shows class-wise calibrated Angular Gap (d) and AVH (e) and calibration (f). Note that the magnitude of the average Angular Gap is more observable than the average AVH across different models. Class-wise model calibration reduces uncertainty in image difficulty estimation.}
\label{fig:uncertainty}
\end{figure}
\paragraph{Multilevel calibration}For calibration, we follow recent papers \cite{guo2017calibration}\cite{nixon2019measuring} and set the initial weights of vector scaling as an identity matrix. We randomly select 10\% of training samples of CIFAR10 for post-training calibration. For ImageNet, we use the validation datasets provided. We optimize the loss functions with LBFGS optimization for 10 epochs with the learning rate set as 0.01. For all experiments, except ImageNet, we report the mean correlation coefficient over 5 different seeds. For ImageNet, we report mean experimental results over 3 different seeds. \\
 \paragraph{Curriculum learning evaluation}Following the fixed easy-to-hard data order as stated in \cite{Bengio2009CurriculumL}, we use the paced learning (PL) setup to fairly compare Angular Gap with other example difficulty measurements. During data loading, we add a fraction of harder examples at the tail of our training data sequence after the current data loader is consumed. Following \cite{wu2020curricula}, we employ ResNet18, linear pacing functions and curriculum learning search grids. Note that, for CIFAR10-H, we use the standard image augmentation and weight decay regularization as stated in \cite{NEURIPS2019_9015} instead of AugMix\cite{hendrycks2019augmix} to ensure consistency. To amplify the effects of image difficulty, we apply cosine learning rate annealing to SGD optmization. As shown in Figure A.7, we have evaluated image difficulty metrics on standard curriculum learning with Weights and Biases\cite{wandb} and Google Cloud Platform. The results are summarized in Table~\ref{tab:main}.
\subsection{Results}
 The results for ImageNetV2 are listed in Table~\ref{tab:imagenetv2} and the results for CIFAR10-H are tabulated as the first columns in Table~\ref{tab:main}. Figure \ref{fig:uncertainty} shows that model calibration has reduced the uncertainty of image difficulty. Our experiments show that simply scaling up models has caused overfitting, and the corresponding image difficulty becomes more uncertain.In Figure \ref{fig:overview}, we have projected the latent features to a unit globe. Easy samples are closer to their class centers and have larger Angular Gap, while difficult samples with negative Angular Gap can be misclassified due to low-resoluted ambiguous content. The 3d visualization reveals more complex angular information, i.e., ambiguous samples related to multiple classes.\\
 We analyze difficulty uncertainty and model capacity using ResNet18, ResNet50 and ResNet101\cite{he2016deep}. In general, Figure \ref{fig:uncertainty} shows that larger CNNs are more poorly calibrated and have shown more uncertainty in image difficulty estimation. Scaling up models will cause overfitting, and the corresponding image difficulty will be less plausible. We also analyze the effects of feature normalization by comparing the training dynamics of AlexNet \cite{krizhevsky2012imagenet}, VGG16 \cite{simonyan2014very} and ResNet50 in the supplementary material as shown Figure A.1 and A.2. The improvements on difficulty estimation may probably come from lower feature norms. By emphasizing angular discrimination, feature normalization and rescaling improve model calibration as shown by the reliability diagrams in Figure A.5. Class-wise calibration further improves the class-wise similarity estimation which is the model confidence in the hyperspherical learning setting.\\
Regarding correlation with human predictions, measured by Spearman's rank and Kendall's Tau, calibrated Angular Gap significantly outperforms other baselines except C-score which is an intensively computational ensembling method. We conclude there are two main reasons. Firstly, the Angular Gap regularizes its difficulty estimation with hyperspherical learning. Secondly, the Angular Gap avoids the hazard of uncertain angular distance by using the largest similarity among other classes, which is powerful when the data point is near the boundary of two classes. Interestingly, Figure A.3 shows that class-wise calibration is similar to human judgements to the classes of CIFAR10-H.  
 Another noteworthy observation is that other difficulty measures show improvement after calibration, suggesting that the standard curriculum learning benefits from better example difficulty estimation. Our results also align with experiments of \cite{kull2019beyond} that show multiclass probabilities can be improved by fine-grained calibration.
\subsection{Domain Adaptation}
\begin{figure}[h]
    \centering
    \includegraphics[width=1\linewidth]{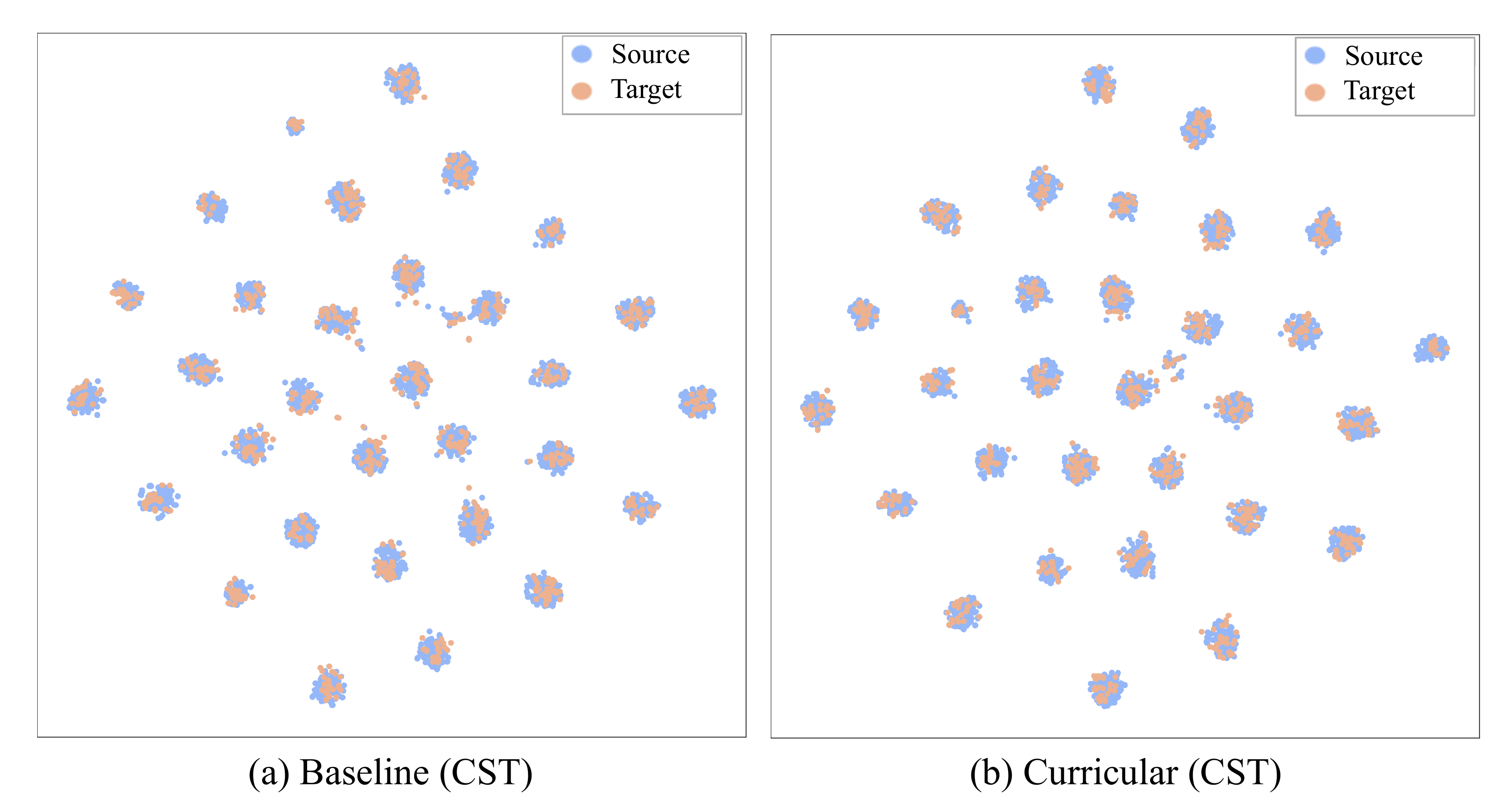}
    \caption{Visualization of features on $\mathbf{ A(source)\rightarrow W(target)}$.}
    \label{fig:tsne}
\end{figure}

We investigate Angular Gap on the domain adaptation tasks for further insights. Following standard protocols of UDA, we use ResNet50 as the backbone for image classification tasks on Office31 and ResNet101 for image classification tasks on VisDA2017. For all methods mentioned above, we project latent features extracted by the backbones to 256d embeddings for discrepancy estimation. For Office31, we use mini-batch stochastic gradient descent (SGD) with an initial learning rate of 0.001, a momentum of 0.9, a batch size of 64, and a weight decay of 5e-4. The pacing function $\lambda$ linearly decreases from 4 to -2 for 100 epochs. For VisDA2017, we set the initial learning rate as 0.0001, a momentum of 0.9, a batchsize of 64, and a weight decay of 5e-4. The pacing function $\lambda$ linearly decreases from 32 to -8 for 100 epochs. For difficulty estimation, we use Adam to finetune the Angular Gap with 80 percent of source data and perform vector scaling calibration with LBFGS optimization on the rest of source data.  The search space is symmetric with $a$ and $b$ chosen from $\{-32, -16,...-2, -1,2,...16,32\}$. We speed up searching with random search and Hyperband\cite{li2017hyperband} algorithms.\\
\begin{table*}
\caption{Accuracy (\%) on Office31. Our Curricular CST method outperforms baselines on six domain adaptation tasks. A$\rightarrow$ W denotes the transformation task from the Amazon domain to Webcam domain, and D$\rightarrow$W denotes the transformation from the Webcam domain to DSLR domain. The best accuracy is indicated in bold, and the second best is underlined.}
  \begin{tabular}{cccccccc}
    \toprule
    Method&A$\rightarrow$ W&D$\rightarrow$ W&W$\rightarrow$ D&
    A$\rightarrow$ D&D$\rightarrow$ A& W$\rightarrow$ A&Avg\\
    \midrule
    ResNet& 68.44$\pm$0.4&96.7$\pm$0.1&99.3$\pm$0.1&68.9$\pm$0.2&62.5$\pm$0.3&60.7$\pm$0.3&76.1\\
    DANN\cite{ganin2015unsupervised} & 84.5$\pm$0.4&96.35$\pm$0.2&99.5$\pm$0.1&80.93$\pm$0.5&69.03$\pm$0.4&68.98$\pm$0.5&83.2\\
    CBST\cite{zou2018unsupervised}&87.8$\pm$0.8&98.5$\pm$0.1&\textbf{100.0$\pm$0.0}&86.5$\pm$1.0&71.2$\pm$0.4&70.9$\pm$0.7&85.8\\
    MSTN\cite{xie2018learning}&91.3$\pm$0.2&98.9$\pm$0.1&\textbf{100.0$\pm$0.0}&90.4$\pm$0.3&72.7$\pm$0.3&65.6$\pm$0.5&86.5\\
        CRST\cite{zou2019confidence}&89.4$\pm$0.7&98.9$\pm$0.4&\textbf{100.0$\pm$0.0}&88.7$\pm$0.8&72.6$\pm$0.7&70.9$\pm$0.5&86.8\\
    DSAN\cite{zhu2020deep}&93.0$\pm$0.4&97.8$\pm$0.2&\textbf{100.0$\pm$0.0}&89.3$\pm$0.7&73.5$\pm$0.5&74.3$\pm$0.4&88.0\\
    CST\cite{liu2021cycle}&95.6$\pm$0.3&98.4$\pm$0.2&\textbf{100.0$\pm$0.0}&95.1$\pm$0.3&77.8$\pm$0.7&78.9$\pm$0.2&91.0\\
    FixBi\cite{na2021fixbi}& \textbf{96.1$\pm$0.2}&\underline{99.3$\pm$0.2}&\textbf{100.0$\pm$0.0}&\textbf{95.0$\pm$0.4}&\underline{78.7$\pm$0.5}&\underline{79.4$\pm$0.3}&\underline{91.4}\\
\midrule
    Curricular DSAN&93.8$\pm$0.2&98.3$\pm$0.1&\textbf{100.0$\pm$0.0}&90.3$\pm$0.5&74.0$\pm$0.3&75.2$\pm$0.4&88.6\\
    Curricular CST&\underline{96.0$\pm$0.1}&\textbf{99.5$\pm$0.2}&\textbf{100.0$\pm$0.0}&\underline{94.9$\pm$0.2}&\textbf{78.9$\pm$0.5}&\textbf{80.4$\pm$0.1}&\textbf{91.6}\\
  \bottomrule
\end{tabular}
\label{tab:office31}
\end{table*}
\begin{table}
\caption{Accuracy (\%) for sythetic-to-real on VisDA2017}
  \begin{tabular}{cc||cc}
    \toprule
    Method & Acc. & Method& Acc.\\
    \midrule
    DANN\cite{ganin2015unsupervised}&55.3&CBST\cite{zou2018unsupervised}&76.4\\
    DAN\cite{long2015learning}&61.1&CRST\cite{zou2019confidence}&78.1\\
    MSTN\cite{xie2018learning}&65.0&FixMatch\cite{sohn2020fixmatch}&76.7\\
    JAN\cite{long2017deep}&65.7&CST\cite{liu2021cycle}&79.9\\
    DSAN\cite{zhu2020deep}&74.8&FixBi\cite{na2021fixbi}&\underline{87.2}\\
    \midrule
    Curricular DSAN&75.4&Curricular CST&\textbf{88.1}\\
    \bottomrule
  \end{tabular}
  \label{tab:visda}
\end{table}
\paragraph{Curricular cycle self-trainig}Table \ref{tab:office31} shows the classification accuracy of our curricular UDA methods on Office31. Table \ref{tab:visda} shows results on {\itshape VisDA-2017}. Curricular CST significantly outperforms state-of-the-art baselines with observable margins, indicating the benefits of the proposed curriculum on domain adaptation tasks. Curricular DSAN has also surpassed several feature alignment methods presented in the left part of the Table \ref{tab:visda}. The optimal curriculum reported by grid search suggests aligning easy samples in the first stage and then focusing on hard samples in the second stage. Note that Curricular CST does not need manually tuning confidence threshold for pseudo-label generation as done with CBST, CRST and FixBi.\\
\paragraph{Curricular domain discrepancy} Figure \ref{fig:sym} shows the dynamics of accuracy (\%) and discrepancy measured by MMD. Curricular CST is able to achieve better training dynamics and final accuracy. There is a noticable flunctuation between 20 to 40 epoch, although sigmoid-shape curriculum provides {\itshape "smooth"} transitions. This indicates the model is able to transit from aligning easy samples to hard sample mining. This aligns with the finding that mining more "informative" samples close to the boundary contributes to better classification results. CST methods have larger MMD than DSAN, but better final performance. We claim that this occurs because neural classifiers can measure discrepancy as logits which is an nonlinear measurement.\\
\paragraph{Feature visualization}
On the task $A \rightarrow W$, we visualize the image embeddings with t-SNE\cite{van2008visualizing} in Figure \ref{fig:tsne}. For CST, both the source and target domain features have formed clusters, but many target samples fall out from their class clusters as outliers. For Curricular CST, although the alignment between source cluster centers and target cluster centers is weaker than CST, target embeddings have successfully formed more compact clusters. As a result, there are less outliers than the baseline.
\\
\paragraph{Limitations}
The proposed methods have potential limitations. Measuring difficulty with Angular Gap inevitably generates computational overheads before curriculum learning. Using hyperspherical learning, Angular Gap based curriculum learning improves model generalization with additional complexity according to \cite{deng2019arcface}.
\section{Conclusions}
In this paper, we propose Angular Gap to address the uncertain image difficulty estimation on a hyperbolic space. We further propose multilevel calibration methods to improve the credibility of the learnt angular metric and look at the calibration problem from hyperspherical learning with geometric interpretations. A curricular cycle self-training method is boosted by the Angular Gap and a curriculum that provides smooth transitions between aligning easy sample and hard sample mining. In our experiments, we show that calibrated Angular Gap is highly correlated with human judgments. On the standard curriculum learning task, the results of Angular Gap are comparable with deep ensembles. We observe the uncertainty of difficulty estimation reduces after calibration. The proposed curriculum results in better optimization of feature discrepancy and significantly improves baselines on the domain adaptation task. Future work can generalize this framework for standard classification tasks and delve into validating model robustness on synthetic data shift with corruption and perturbation.
\bibliographystyle{ACM-Reference-Format}
\bibliography{sample-base}


\begin{thebibliography}{47}


\ifx \showCODEN    \undefined \def \showCODEN     #1{\unskip}     \fi
\ifx \showDOI      \undefined \def \showDOI       #1{#1}\fi
\ifx \showISBNx    \undefined \def \showISBNx     #1{\unskip}     \fi
\ifx \showISBNxiii \undefined \def \showISBNxiii  #1{\unskip}     \fi
\ifx \showISSN     \undefined \def \showISSN      #1{\unskip}     \fi
\ifx \showLCCN     \undefined \def \showLCCN      #1{\unskip}     \fi
\ifx \shownote     \undefined \def \shownote      #1{#1}          \fi
\ifx \showarticletitle \undefined \def \showarticletitle #1{#1}   \fi
\ifx \showURL      \undefined \def \showURL       {\relax}        \fi
\providecommand\bibfield[2]{#2}
\providecommand\bibinfo[2]{#2}
\providecommand\natexlab[1]{#1}
\providecommand\showeprint[2][]{arXiv:#2}

\bibitem[Baldock et~al\mbox{.}(2021)]%
        {baldock2021deep}
\bibfield{author}{\bibinfo{person}{Robert Baldock}, \bibinfo{person}{Hartmut
  Maennel}, {and} \bibinfo{person}{Behnam Neyshabur}.}
  \bibinfo{year}{2021}\natexlab{}.
\newblock \showarticletitle{Deep learning through the lens of example
  difficulty}.
\newblock \bibinfo{journal}{\emph{Advances in Neural Information Processing
  Systems}}  \bibinfo{volume}{34} (\bibinfo{year}{2021}).
\newblock


\bibitem[Battleday et~al\mbox{.}(2020)]%
        {Battleday2020CapturingHC}
\bibfield{author}{\bibinfo{person}{R. Battleday}, \bibinfo{person}{Joshua~C.
  Peterson}, {and} \bibinfo{person}{T. Griffiths}.}
  \bibinfo{year}{2020}\natexlab{}.
\newblock \showarticletitle{Capturing human categorization of natural images by
  combining deep networks and cognitive models}.
\newblock \bibinfo{journal}{\emph{Nature Communications}}  \bibinfo{volume}{11}
  (\bibinfo{year}{2020}).
\newblock


\bibitem[Bengio et~al\mbox{.}(2009)]%
        {Bengio2009CurriculumL}
\bibfield{author}{\bibinfo{person}{Yoshua Bengio}, \bibinfo{person}{J.
  Louradour}, \bibinfo{person}{Ronan Collobert}, {and} \bibinfo{person}{J.
  Weston}.} \bibinfo{year}{2009}\natexlab{}.
\newblock \showarticletitle{Curriculum learning}. In
  \bibinfo{booktitle}{\emph{ICML '09}}.
\newblock


\bibitem[Biewald(2020)]%
        {wandb}
\bibfield{author}{\bibinfo{person}{Lukas Biewald}.}
  \bibinfo{year}{2020}\natexlab{}.
\newblock \bibinfo{title}{Experiment Tracking with Weights and Biases}.
\newblock
\newblock
\urldef\tempurl%
\url{https://www.wandb.com/}
\showURL{%
\tempurl}
\newblock
\shownote{Software available from wandb.com}.


\bibitem[Chen et~al\mbox{.}(2020b)]%
        {chen2020angular}
\bibfield{author}{\bibinfo{person}{Beidi Chen}, \bibinfo{person}{Weiyang Liu},
  \bibinfo{person}{Zhiding Yu}, \bibinfo{person}{Jan Kautz},
  \bibinfo{person}{Anshumali Shrivastava}, \bibinfo{person}{Animesh Garg},
  {and} \bibinfo{person}{Animashree Anandkumar}.}
  \bibinfo{year}{2020}\natexlab{b}.
\newblock \showarticletitle{Angular visual hardness}. In
  \bibinfo{booktitle}{\emph{International Conference on Machine Learning}}.
  PMLR, \bibinfo{pages}{1637--1648}.
\newblock


\bibitem[Chen et~al\mbox{.}(2020a)]%
        {chen2020simple}
\bibfield{author}{\bibinfo{person}{Ting Chen}, \bibinfo{person}{Simon
  Kornblith}, \bibinfo{person}{Mohammad Norouzi}, {and}
  \bibinfo{person}{Geoffrey Hinton}.} \bibinfo{year}{2020}\natexlab{a}.
\newblock \showarticletitle{A simple framework for contrastive learning of
  visual representations}. In \bibinfo{booktitle}{\emph{International
  conference on machine learning}}. PMLR, \bibinfo{pages}{1597--1607}.
\newblock


\bibitem[Deng et~al\mbox{.}(2019)]%
        {deng2019arcface}
\bibfield{author}{\bibinfo{person}{Jiankang Deng}, \bibinfo{person}{Jia Guo},
  \bibinfo{person}{Niannan Xue}, {and} \bibinfo{person}{Stefanos Zafeiriou}.}
  \bibinfo{year}{2019}\natexlab{}.
\newblock \showarticletitle{Arcface: Additive angular margin loss for deep face
  recognition}. In \bibinfo{booktitle}{\emph{Proceedings of the IEEE/CVF
  Conference on Computer Vision and Pattern Recognition}}.
  \bibinfo{pages}{4690--4699}.
\newblock


\bibitem[Feng et~al\mbox{.}(2020)]%
        {feng2020language}
\bibfield{author}{\bibinfo{person}{Fangxiaoyu Feng}, \bibinfo{person}{Yinfei
  Yang}, \bibinfo{person}{Daniel Cer}, \bibinfo{person}{Naveen Arivazhagan},
  {and} \bibinfo{person}{Wei Wang}.} \bibinfo{year}{2020}\natexlab{}.
\newblock \showarticletitle{Language-agnostic bert sentence embedding}.
\newblock \bibinfo{journal}{\emph{arXiv preprint arXiv:2007.01852}}
  (\bibinfo{year}{2020}).
\newblock


\bibitem[French(1999)]%
        {french1999catastrophic}
\bibfield{author}{\bibinfo{person}{Robert~M French}.}
  \bibinfo{year}{1999}\natexlab{}.
\newblock \showarticletitle{Catastrophic forgetting in connectionist networks}.
\newblock \bibinfo{journal}{\emph{Trends in cognitive sciences}}
  \bibinfo{volume}{3}, \bibinfo{number}{4} (\bibinfo{year}{1999}),
  \bibinfo{pages}{128--135}.
\newblock


\bibitem[Gal and Ghahramani(2016)]%
        {gal2016dropout}
\bibfield{author}{\bibinfo{person}{Yarin Gal} {and} \bibinfo{person}{Zoubin
  Ghahramani}.} \bibinfo{year}{2016}\natexlab{}.
\newblock \showarticletitle{Dropout as a bayesian approximation: Representing
  model uncertainty in deep learning}. In
  \bibinfo{booktitle}{\emph{international conference on machine learning}}.
  PMLR, \bibinfo{pages}{1050--1059}.
\newblock


\bibitem[Ganin and Lempitsky(2015)]%
        {ganin2015unsupervised}
\bibfield{author}{\bibinfo{person}{Yaroslav Ganin} {and}
  \bibinfo{person}{Victor Lempitsky}.} \bibinfo{year}{2015}\natexlab{}.
\newblock \showarticletitle{Unsupervised domain adaptation by backpropagation}.
  In \bibinfo{booktitle}{\emph{International conference on machine learning}}.
  PMLR, \bibinfo{pages}{1180--1189}.
\newblock


\bibitem[Ghifary et~al\mbox{.}(2014)]%
        {ghifary2014domain}
\bibfield{author}{\bibinfo{person}{Muhammad Ghifary},
  \bibinfo{person}{W~Bastiaan Kleijn}, {and} \bibinfo{person}{Mengjie Zhang}.}
  \bibinfo{year}{2014}\natexlab{}.
\newblock \showarticletitle{Domain adaptive neural networks for object
  recognition}. In \bibinfo{booktitle}{\emph{Pacific Rim international
  conference on artificial intelligence}}. Springer, \bibinfo{pages}{898--904}.
\newblock


\bibitem[Guo et~al\mbox{.}(2017)]%
        {guo2017calibration}
\bibfield{author}{\bibinfo{person}{Chuan Guo}, \bibinfo{person}{Geoff Pleiss},
  \bibinfo{person}{Yu Sun}, {and} \bibinfo{person}{Kilian~Q Weinberger}.}
  \bibinfo{year}{2017}\natexlab{}.
\newblock \showarticletitle{On calibration of modern neural networks}. In
  \bibinfo{booktitle}{\emph{International Conference on Machine Learning}}.
  PMLR, \bibinfo{pages}{1321--1330}.
\newblock


\bibitem[He et~al\mbox{.}(2016)]%
        {he2016deep}
\bibfield{author}{\bibinfo{person}{Kaiming He}, \bibinfo{person}{Xiangyu
  Zhang}, \bibinfo{person}{Shaoqing Ren}, {and} \bibinfo{person}{Jian Sun}.}
  \bibinfo{year}{2016}\natexlab{}.
\newblock \showarticletitle{Deep residual learning for image recognition}. In
  \bibinfo{booktitle}{\emph{Proceedings of the IEEE conference on computer
  vision and pattern recognition}}. \bibinfo{pages}{770--778}.
\newblock


\bibitem[Hendrycks et~al\mbox{.}(2019)]%
        {hendrycks2019augmix}
\bibfield{author}{\bibinfo{person}{Dan Hendrycks}, \bibinfo{person}{Norman Mu},
  \bibinfo{person}{Ekin~D Cubuk}, \bibinfo{person}{Barret Zoph},
  \bibinfo{person}{Justin Gilmer}, {and} \bibinfo{person}{Balaji
  Lakshminarayanan}.} \bibinfo{year}{2019}\natexlab{}.
\newblock \showarticletitle{Augmix: A simple data processing method to improve
  robustness and uncertainty}.
\newblock \bibinfo{journal}{\emph{arXiv preprint arXiv:1912.02781}}
  (\bibinfo{year}{2019}).
\newblock


\bibitem[Islam et~al\mbox{.}(2021)]%
        {islam2021class}
\bibfield{author}{\bibinfo{person}{Mobarakol Islam},
  \bibinfo{person}{Lalithkumar Seenivasan}, \bibinfo{person}{Hongliang Ren},
  {and} \bibinfo{person}{Ben Glocker}.} \bibinfo{year}{2021}\natexlab{}.
\newblock \showarticletitle{Class-distribution-aware calibration for
  long-tailed visual recognition}.
\newblock \bibinfo{journal}{\emph{arXiv preprint arXiv:2109.05263}}
  (\bibinfo{year}{2021}).
\newblock


\bibitem[Jiang et~al\mbox{.}(2018)]%
        {jiang2018mentornet}
\bibfield{author}{\bibinfo{person}{Lu Jiang}, \bibinfo{person}{Zhengyuan Zhou},
  \bibinfo{person}{Thomas Leung}, \bibinfo{person}{Li-Jia Li}, {and}
  \bibinfo{person}{Li Fei-Fei}.} \bibinfo{year}{2018}\natexlab{}.
\newblock \showarticletitle{Mentornet: Learning data-driven curriculum for very
  deep neural networks on corrupted labels}. In
  \bibinfo{booktitle}{\emph{International Conference on Machine Learning}}.
  PMLR, \bibinfo{pages}{2304--2313}.
\newblock


\bibitem[Jiang et~al\mbox{.}(2020)]%
        {jiang2020characterizing}
\bibfield{author}{\bibinfo{person}{Ziheng Jiang}, \bibinfo{person}{Chiyuan
  Zhang}, \bibinfo{person}{Kunal Talwar}, {and} \bibinfo{person}{Michael~C
  Mozer}.} \bibinfo{year}{2020}\natexlab{}.
\newblock \showarticletitle{Characterizing structural regularities of labeled
  data in overparameterized models}.
\newblock \bibinfo{journal}{\emph{arXiv preprint arXiv:2002.03206}}
  (\bibinfo{year}{2020}).
\newblock


\bibitem[Krizhevsky et~al\mbox{.}(2009)]%
        {krizhevsky2009learning}
\bibfield{author}{\bibinfo{person}{Alex Krizhevsky}, \bibinfo{person}{Geoffrey
  Hinton}, {et~al\mbox{.}}} \bibinfo{year}{2009}\natexlab{}.
\newblock \showarticletitle{Learning multiple layers of features from tiny
  images}.
\newblock  (\bibinfo{year}{2009}).
\newblock


\bibitem[Krizhevsky et~al\mbox{.}(2012)]%
        {krizhevsky2012imagenet}
\bibfield{author}{\bibinfo{person}{Alex Krizhevsky}, \bibinfo{person}{Ilya
  Sutskever}, {and} \bibinfo{person}{Geoffrey~E Hinton}.}
  \bibinfo{year}{2012}\natexlab{}.
\newblock \showarticletitle{Imagenet classification with deep convolutional
  neural networks}.
\newblock \bibinfo{journal}{\emph{Advances in neural information processing
  systems}}  \bibinfo{volume}{25} (\bibinfo{year}{2012}).
\newblock


\bibitem[Kull et~al\mbox{.}(2019)]%
        {kull2019beyond}
\bibfield{author}{\bibinfo{person}{Meelis Kull}, \bibinfo{person}{Miquel
  Perello~Nieto}, \bibinfo{person}{Markus K{\"a}ngsepp}, \bibinfo{person}{Telmo
  Silva~Filho}, \bibinfo{person}{Hao Song}, {and} \bibinfo{person}{Peter
  Flach}.} \bibinfo{year}{2019}\natexlab{}.
\newblock \showarticletitle{Beyond temperature scaling: Obtaining
  well-calibrated multi-class probabilities with dirichlet calibration}.
\newblock \bibinfo{journal}{\emph{Advances in neural information processing
  systems}}  \bibinfo{volume}{32} (\bibinfo{year}{2019}).
\newblock


\bibitem[Kumar et~al\mbox{.}(2010)]%
        {kumar2010self}
\bibfield{author}{\bibinfo{person}{M Kumar}, \bibinfo{person}{Benjamin Packer},
  {and} \bibinfo{person}{Daphne Koller}.} \bibinfo{year}{2010}\natexlab{}.
\newblock \showarticletitle{Self-paced learning for latent variable models}.
\newblock \bibinfo{journal}{\emph{Advances in neural information processing
  systems}}  \bibinfo{volume}{23} (\bibinfo{year}{2010}).
\newblock


\bibitem[Lakshminarayanan et~al\mbox{.}(2017)]%
        {lakshminarayanan2017simple}
\bibfield{author}{\bibinfo{person}{Balaji Lakshminarayanan},
  \bibinfo{person}{Alexander Pritzel}, {and} \bibinfo{person}{Charles
  Blundell}.} \bibinfo{year}{2017}\natexlab{}.
\newblock \showarticletitle{Simple and scalable predictive uncertainty
  estimation using deep ensembles}.
\newblock \bibinfo{journal}{\emph{Advances in neural information processing
  systems}}  \bibinfo{volume}{30} (\bibinfo{year}{2017}).
\newblock


\bibitem[Li et~al\mbox{.}(2017)]%
        {li2017hyperband}
\bibfield{author}{\bibinfo{person}{Lisha Li}, \bibinfo{person}{Kevin Jamieson},
  \bibinfo{person}{Giulia DeSalvo}, \bibinfo{person}{Afshin Rostamizadeh},
  {and} \bibinfo{person}{Ameet Talwalkar}.} \bibinfo{year}{2017}\natexlab{}.
\newblock \showarticletitle{Hyperband: A novel bandit-based approach to
  hyperparameter optimization}.
\newblock \bibinfo{journal}{\emph{The Journal of Machine Learning Research}}
  \bibinfo{volume}{18}, \bibinfo{number}{1} (\bibinfo{year}{2017}),
  \bibinfo{pages}{6765--6816}.
\newblock


\bibitem[Liu et~al\mbox{.}(2021)]%
        {liu2021cycle}
\bibfield{author}{\bibinfo{person}{Hong Liu}, \bibinfo{person}{Jianmin Wang},
  {and} \bibinfo{person}{Mingsheng Long}.} \bibinfo{year}{2021}\natexlab{}.
\newblock \showarticletitle{Cycle self-training for domain adaptation}.
\newblock \bibinfo{journal}{\emph{Advances in Neural Information Processing
  Systems}}  \bibinfo{volume}{34} (\bibinfo{year}{2021}).
\newblock


\bibitem[Liu et~al\mbox{.}(2017)]%
        {liu2017deep}
\bibfield{author}{\bibinfo{person}{Weiyang Liu}, \bibinfo{person}{Yan-Ming
  Zhang}, \bibinfo{person}{Xingguo Li}, \bibinfo{person}{Zhiding Yu},
  \bibinfo{person}{Bo Dai}, \bibinfo{person}{Tuo Zhao}, {and}
  \bibinfo{person}{Le Song}.} \bibinfo{year}{2017}\natexlab{}.
\newblock \showarticletitle{Deep hyperspherical learning}.
\newblock \bibinfo{journal}{\emph{Advances in neural information processing
  systems}}  \bibinfo{volume}{30} (\bibinfo{year}{2017}).
\newblock


\bibitem[Long et~al\mbox{.}(2015)]%
        {long2015learning}
\bibfield{author}{\bibinfo{person}{Mingsheng Long}, \bibinfo{person}{Yue Cao},
  \bibinfo{person}{Jianmin Wang}, {and} \bibinfo{person}{Michael Jordan}.}
  \bibinfo{year}{2015}\natexlab{}.
\newblock \showarticletitle{Learning transferable features with deep adaptation
  networks}. In \bibinfo{booktitle}{\emph{International conference on machine
  learning}}. PMLR, \bibinfo{pages}{97--105}.
\newblock


\bibitem[Long et~al\mbox{.}(2017)]%
        {long2017deep}
\bibfield{author}{\bibinfo{person}{Mingsheng Long}, \bibinfo{person}{Han Zhu},
  \bibinfo{person}{Jianmin Wang}, {and} \bibinfo{person}{Michael~I Jordan}.}
  \bibinfo{year}{2017}\natexlab{}.
\newblock \showarticletitle{Deep transfer learning with joint adaptation
  networks}. In \bibinfo{booktitle}{\emph{International conference on machine
  learning}}. PMLR, \bibinfo{pages}{2208--2217}.
\newblock


\bibitem[Na et~al\mbox{.}(2021)]%
        {na2021fixbi}
\bibfield{author}{\bibinfo{person}{Jaemin Na}, \bibinfo{person}{Heechul Jung},
  \bibinfo{person}{Hyung~Jin Chang}, {and} \bibinfo{person}{Wonjun Hwang}.}
  \bibinfo{year}{2021}\natexlab{}.
\newblock \showarticletitle{Fixbi: Bridging domain spaces for unsupervised
  domain adaptation}. In \bibinfo{booktitle}{\emph{Proceedings of the IEEE/CVF
  Conference on Computer Vision and Pattern Recognition}}.
  \bibinfo{pages}{1094--1103}.
\newblock


\bibitem[Nixon et~al\mbox{.}(2019)]%
        {nixon2019measuring}
\bibfield{author}{\bibinfo{person}{Jeremy Nixon}, \bibinfo{person}{Michael~W
  Dusenberry}, \bibinfo{person}{Linchuan Zhang}, \bibinfo{person}{Ghassen
  Jerfel}, {and} \bibinfo{person}{Dustin Tran}.}
  \bibinfo{year}{2019}\natexlab{}.
\newblock \showarticletitle{Measuring Calibration in Deep Learning.}. In
  \bibinfo{booktitle}{\emph{CVPR Workshops}}, Vol.~\bibinfo{volume}{2}.
\newblock


\bibitem[Paszke and Gross(2019)]%
        {NEURIPS2019_9015}
\bibfield{author}{\bibinfo{person}{Adam Paszke} {and} \bibinfo{person}{Gross}.}
  \bibinfo{year}{2019}\natexlab{}.
\newblock \showarticletitle{PyTorch: An Imperative Style, High-Performance Deep
  Learning Library}.
\newblock In \bibinfo{booktitle}{\emph{Advances in Neural Information
  Processing Systems 32}}. \bibinfo{publisher}{Curran Associates, Inc.},
  \bibinfo{pages}{8024--8035}.
\newblock
\urldef\tempurl%
\url{https://github.com/pytorch/examples/blob/master/imagenet/main.py}
\showURL{%
\tempurl}


\bibitem[Peng et~al\mbox{.}(2017)]%
        {peng2017visda}
\bibfield{author}{\bibinfo{person}{Xingchao Peng}, \bibinfo{person}{Ben Usman},
  \bibinfo{person}{Neela Kaushik}, \bibinfo{person}{Judy Hoffman},
  \bibinfo{person}{Dequan Wang}, {and} \bibinfo{person}{Kate Saenko}.}
  \bibinfo{year}{2017}\natexlab{}.
\newblock \showarticletitle{Visda: The visual domain adaptation challenge}.
\newblock \bibinfo{journal}{\emph{arXiv preprint arXiv:1710.06924}}
  (\bibinfo{year}{2017}).
\newblock


\bibitem[Platt et~al\mbox{.}(1999)]%
        {platt1999probabilistic}
\bibfield{author}{\bibinfo{person}{John Platt} {et~al\mbox{.}}}
  \bibinfo{year}{1999}\natexlab{}.
\newblock \showarticletitle{Probabilistic outputs for support vector machines
  and comparisons to regularized likelihood methods}.
\newblock \bibinfo{journal}{\emph{Advances in large margin classifiers}}
  \bibinfo{volume}{10}, \bibinfo{number}{3} (\bibinfo{year}{1999}),
  \bibinfo{pages}{61--74}.
\newblock


\bibitem[Recht et~al\mbox{.}(2019)]%
        {recht2019imagenet}
\bibfield{author}{\bibinfo{person}{Benjamin Recht}, \bibinfo{person}{Rebecca
  Roelofs}, \bibinfo{person}{Ludwig Schmidt}, {and} \bibinfo{person}{Vaishaal
  Shankar}.} \bibinfo{year}{2019}\natexlab{}.
\newblock \showarticletitle{Do imagenet classifiers generalize to imagenet?}.
  In \bibinfo{booktitle}{\emph{International Conference on Machine Learning}}.
  PMLR, \bibinfo{pages}{5389--5400}.
\newblock


\bibitem[Saenko et~al\mbox{.}(2010)]%
        {saenko2010adapting}
\bibfield{author}{\bibinfo{person}{Kate Saenko}, \bibinfo{person}{Brian Kulis},
  \bibinfo{person}{Mario Fritz}, {and} \bibinfo{person}{Trevor Darrell}.}
  \bibinfo{year}{2010}\natexlab{}.
\newblock \showarticletitle{Adapting visual category models to new domains}. In
  \bibinfo{booktitle}{\emph{European conference on computer vision}}. Springer,
  \bibinfo{pages}{213--226}.
\newblock


\bibitem[Shu et~al\mbox{.}(2019)]%
        {shu2019transferable}
\bibfield{author}{\bibinfo{person}{Yang Shu}, \bibinfo{person}{Zhangjie Cao},
  \bibinfo{person}{Mingsheng Long}, {and} \bibinfo{person}{Jianmin Wang}.}
  \bibinfo{year}{2019}\natexlab{}.
\newblock \showarticletitle{Transferable curriculum for weakly-supervised
  domain adaptation}. In \bibinfo{booktitle}{\emph{Proceedings of the AAAI
  Conference on Artificial Intelligence}}, Vol.~\bibinfo{volume}{33}.
  \bibinfo{pages}{4951--4958}.
\newblock


\bibitem[Simonyan and Zisserman(2014)]%
        {simonyan2014very}
\bibfield{author}{\bibinfo{person}{Karen Simonyan} {and}
  \bibinfo{person}{Andrew Zisserman}.} \bibinfo{year}{2014}\natexlab{}.
\newblock \showarticletitle{Very deep convolutional networks for large-scale
  image recognition}.
\newblock \bibinfo{journal}{\emph{arXiv preprint arXiv:1409.1556}}
  (\bibinfo{year}{2014}).
\newblock


\bibitem[Sohn et~al\mbox{.}(2020)]%
        {sohn2020fixmatch}
\bibfield{author}{\bibinfo{person}{Kihyuk Sohn}, \bibinfo{person}{David
  Berthelot}, \bibinfo{person}{Nicholas Carlini}, \bibinfo{person}{Zizhao
  Zhang}, \bibinfo{person}{Han Zhang}, \bibinfo{person}{Colin~A Raffel},
  \bibinfo{person}{Ekin~Dogus Cubuk}, \bibinfo{person}{Alexey Kurakin}, {and}
  \bibinfo{person}{Chun-Liang Li}.} \bibinfo{year}{2020}\natexlab{}.
\newblock \showarticletitle{Fixmatch: Simplifying semi-supervised learning with
  consistency and confidence}.
\newblock \bibinfo{journal}{\emph{Advances in Neural Information Processing
  Systems}}  \bibinfo{volume}{33} (\bibinfo{year}{2020}),
  \bibinfo{pages}{596--608}.
\newblock


\bibitem[Toneva et~al\mbox{.}(2019)]%
        {Toneva2019AnES}
\bibfield{author}{\bibinfo{person}{Mariya Toneva}, \bibinfo{person}{Alessandro
  Sordoni}, \bibinfo{person}{R{\'e}mi~Tachet des Combes}, \bibinfo{person}{Adam
  Trischler}, \bibinfo{person}{Yoshua Bengio}, {and}
  \bibinfo{person}{Geoffrey~J. Gordon}.} \bibinfo{year}{2019}\natexlab{}.
\newblock \showarticletitle{An Empirical Study of Example Forgetting during
  Deep Neural Network Learning}.
\newblock \bibinfo{journal}{\emph{ArXiv}}  \bibinfo{volume}{abs/1812.05159}
  (\bibinfo{year}{2019}).
\newblock


\bibitem[Van~der Maaten and Hinton(2008)]%
        {van2008visualizing}
\bibfield{author}{\bibinfo{person}{Laurens Van~der Maaten} {and}
  \bibinfo{person}{Geoffrey Hinton}.} \bibinfo{year}{2008}\natexlab{}.
\newblock \showarticletitle{Visualizing data using t-SNE.}
\newblock \bibinfo{journal}{\emph{Journal of machine learning research}}
  \bibinfo{volume}{9}, \bibinfo{number}{11} (\bibinfo{year}{2008}).
\newblock


\bibitem[Wang et~al\mbox{.}(2018)]%
        {wang2018cosface}
\bibfield{author}{\bibinfo{person}{Hao Wang}, \bibinfo{person}{Yitong Wang},
  \bibinfo{person}{Zheng Zhou}, \bibinfo{person}{Xing Ji},
  \bibinfo{person}{Dihong Gong}, \bibinfo{person}{Jingchao Zhou},
  \bibinfo{person}{Zhifeng Li}, {and} \bibinfo{person}{Wei Liu}.}
  \bibinfo{year}{2018}\natexlab{}.
\newblock \showarticletitle{Cosface: Large margin cosine loss for deep face
  recognition}. In \bibinfo{booktitle}{\emph{Proceedings of the IEEE conference
  on computer vision and pattern recognition}}. \bibinfo{pages}{5265--5274}.
\newblock


\bibitem[Wu et~al\mbox{.}(2020)]%
        {wu2020curricula}
\bibfield{author}{\bibinfo{person}{Xiaoxia Wu}, \bibinfo{person}{Ethan Dyer},
  {and} \bibinfo{person}{Behnam Neyshabur}.} \bibinfo{year}{2020}\natexlab{}.
\newblock \showarticletitle{When Do Curricula Work?}
\newblock \bibinfo{journal}{\emph{arXiv preprint arXiv:2012.03107}}
  (\bibinfo{year}{2020}).
\newblock


\bibitem[Xie et~al\mbox{.}(2018)]%
        {xie2018learning}
\bibfield{author}{\bibinfo{person}{Shaoan Xie}, \bibinfo{person}{Zibin Zheng},
  \bibinfo{person}{Liang Chen}, {and} \bibinfo{person}{Chuan Chen}.}
  \bibinfo{year}{2018}\natexlab{}.
\newblock \showarticletitle{Learning semantic representations for unsupervised
  domain adaptation}. In \bibinfo{booktitle}{\emph{International conference on
  machine learning}}. PMLR, \bibinfo{pages}{5423--5432}.
\newblock


\bibitem[Zhang et~al\mbox{.}(2017)]%
        {zhang2017curriculum}
\bibfield{author}{\bibinfo{person}{Yang Zhang}, \bibinfo{person}{Philip David},
  {and} \bibinfo{person}{Boqing Gong}.} \bibinfo{year}{2017}\natexlab{}.
\newblock \showarticletitle{Curriculum domain adaptation for semantic
  segmentation of urban scenes}. In \bibinfo{booktitle}{\emph{Proceedings of
  the IEEE international conference on computer vision}}.
  \bibinfo{pages}{2020--2030}.
\newblock


\bibitem[Zhu et~al\mbox{.}(2020)]%
        {zhu2020deep}
\bibfield{author}{\bibinfo{person}{Yongchun Zhu}, \bibinfo{person}{Fuzhen
  Zhuang}, \bibinfo{person}{Jindong Wang}, \bibinfo{person}{Guolin Ke},
  \bibinfo{person}{Jingwu Chen}, \bibinfo{person}{Jiang Bian},
  \bibinfo{person}{Hui Xiong}, {and} \bibinfo{person}{Qing He}.}
  \bibinfo{year}{2020}\natexlab{}.
\newblock \showarticletitle{Deep subdomain adaptation network for image
  classification}.
\newblock \bibinfo{journal}{\emph{IEEE transactions on neural networks and
  learning systems}} \bibinfo{volume}{32}, \bibinfo{number}{4}
  (\bibinfo{year}{2020}), \bibinfo{pages}{1713--1722}.
\newblock


\bibitem[Zou et~al\mbox{.}(2018)]%
        {zou2018unsupervised}
\bibfield{author}{\bibinfo{person}{Yang Zou}, \bibinfo{person}{Zhiding Yu},
  \bibinfo{person}{BVK Kumar}, {and} \bibinfo{person}{Jinsong Wang}.}
  \bibinfo{year}{2018}\natexlab{}.
\newblock \showarticletitle{Unsupervised domain adaptation for semantic
  segmentation via class-balanced self-training}. In
  \bibinfo{booktitle}{\emph{Proceedings of the European conference on computer
  vision (ECCV)}}. \bibinfo{pages}{289--305}.
\newblock


\bibitem[Zou et~al\mbox{.}(2019)]%
        {zou2019confidence}
\bibfield{author}{\bibinfo{person}{Yang Zou}, \bibinfo{person}{Zhiding Yu},
  \bibinfo{person}{Xiaofeng Liu}, \bibinfo{person}{BVK Kumar}, {and}
  \bibinfo{person}{Jinsong Wang}.} \bibinfo{year}{2019}\natexlab{}.
\newblock \showarticletitle{Confidence regularized self-training}. In
  \bibinfo{booktitle}{\emph{Proceedings of the IEEE/CVF International
  Conference on Computer Vision}}. \bibinfo{pages}{5982--5991}.
\newblock


\end{thebibliography}
\twocolumn[
\centering
\textbf{\Huge Angular Gap: reducing the uncertainty of image difficulty through model calibration\\
Supplementary material
} \\
\vspace{1.0em}
] 
\setcounter{equation}{0}
\setcounter{figure}{0}
\setcounter{table}{0}
\setcounter{page}{1}
\makeatletter
\renewcommand{\theequation}{A.\arabic{equation}}
\renewcommand{\thefigure}{A.\arabic{figure}}
\renewcommand{\bibnumfmt}[1]{[A.#1]}
\renewcommand{\citenumfont}[1]{A.#1}
\appendix
\section{Calibrated Angular Gap}
In this section, we give additional information for Angular Gap and multilevel calibration. First, we clarify the difference between confidence and Angular Gap. Then we show some examples of feature norm plots, class-wise calibration maps and reliability diagrams that can help to understand calibrated Angular Gap.
\subsection{Model confidence and Angular Gap}
In the supervised multi-class classification with neural networks, model confidence refers to the outputs from the softmax layer. Given a class prediction $\hat{y}_i$ of softmax probabilities $\mathbf{y}_i\in\mathbb{R}^C$, the confidence can often be computed as follows 
\begin{equation}\label{eq:confidence}
\mathbb{P}(\hat{y}_i|\mathbf{x}_{i}, \mathbf{W}, \mathbf{b})=\frac{\exp(\mathbf{w}_{\hat{y}_i}^{T} \mathbf{x}_{i}+b_{\hat{y}_i)}}{ \sum_{k=1}^{C} \exp(\mathbf{w}_{k}^{T} \mathbf{x}_{i}+b_{k})}.
\end{equation}
By contrast, Angular Gap, defined in Equation \ref{eq:definition}, uses the cosine similarities between feature vectors and class vectors before the softmax operation. Although the similarity can be considered as a nonlinear version of confidence, we focus on ascertaining the values of these similarities for individual samples during hyperspherical learning. These generalized similarities are enforced by feature normalization and removing bias from the classification layer. To connect Angular Gap with confidence, we write the normalized softmax loss as
\begin{align}
\min_{\mathbf{W, x_{i}}}L_{NSL} &= - \frac{1}{N}\sum_{i=1}^N\log \mathbb{P}(y_i|\mathbf{x}_{i}, \mathbf{W}, s),\\
&= - \frac{1}{N}\sum_{i=1}^N\log \frac{\exp( s\cdot(\mathbf{w}_{y_i}^{T} \mathbf{x}_{i}/\left\|{\mathbf{w}_k}\right\|\left\|{\mathbf x_i}\right\|))}{ \sum_{k=1}^{C} \exp(s\cdot (\mathbf{w}_{k}^{T} \mathbf{x}_{i}/\left\|{\mathbf{w}_k}\right\|\left\|{\mathbf x_i}\right\|))}.
\end{align}
\begin{figure}[!hbp]
\centering
\includegraphics[width=\linewidth]{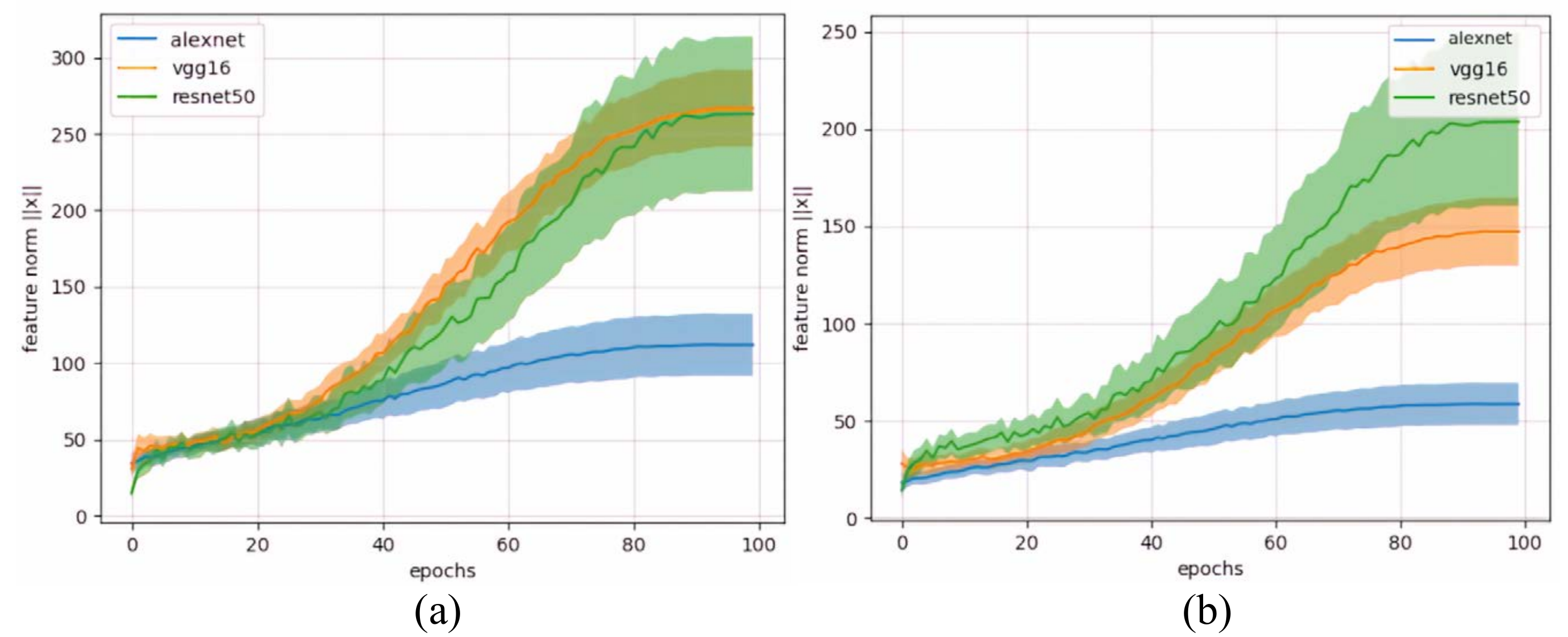}
\caption{Feature norm $\left \|x\right \|$ on CIFAR10-H with shadows represent the standard deviation. (a) Training with cross-entropy loss. (b) Training with normalized softmax loss (NSL). Uncertainty increases as model capacity increases.}
\label{fig:feature_norm_cifar10h}
\end{figure}
\subsection{Feature norm}
\begin{figure}[]
\centering
\includegraphics[width=\linewidth]{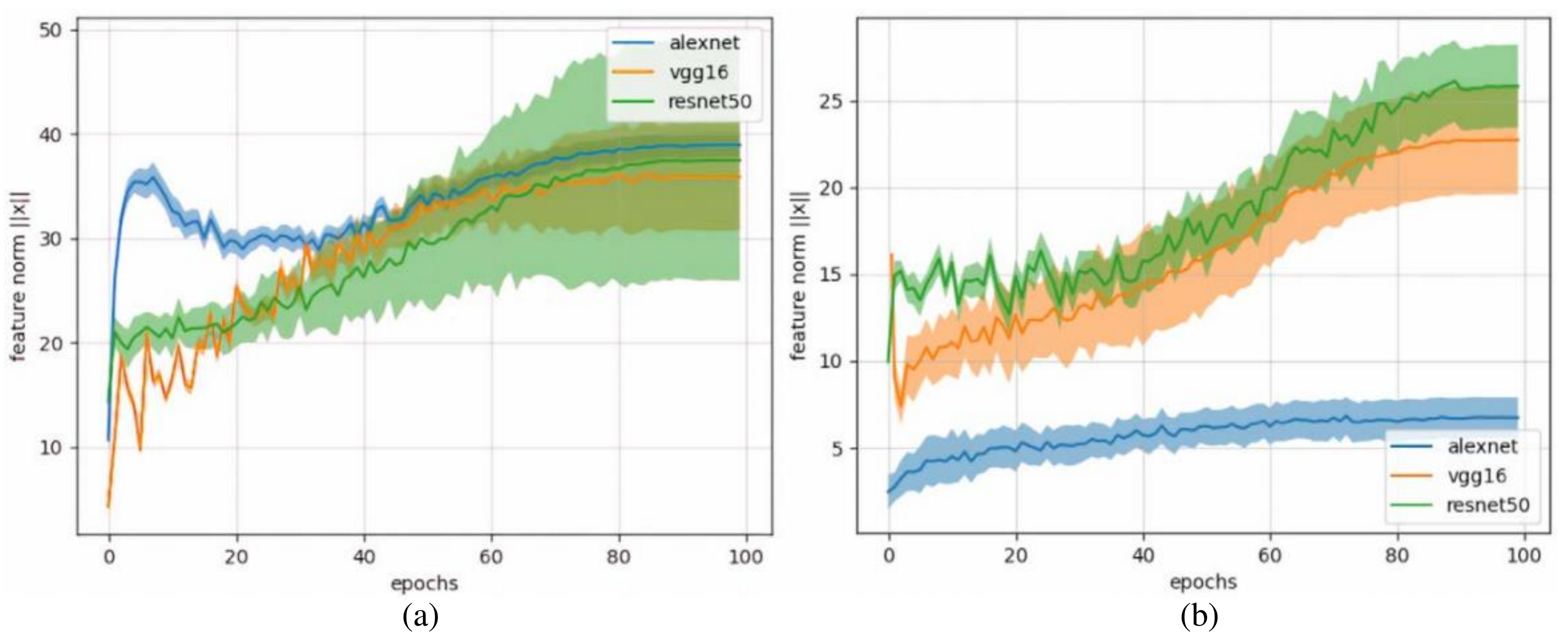}
\caption{Feature norm $\left \|x\right \|$ on ILVRC 2012. (a) Training with cross-entropy loss. (b) Training with NSL.}
\label{fig:feature_norm_imagenetv2}
\end{figure}

Figure \ref{fig:feature_norm_cifar10h} and Figure \ref{fig:feature_norm_imagenetv2} show the dynamics of feature norm trained on CIFAR10 with cross-entropy loss and normalized softmax loss (NSL) respectively. For CIFAR10-H, the mean and standard deviation are reported over five random seeds, while for ImageNetV2 the statistics are computed over three random seeds. The feature norms slowly diverge as the negative log likelihood minimizes. This indicates the necessity of model calibration when using the probabilities output by a single neural classifier. Although training with 
NSL makes feature norm smaller, the magnitude of uncertainty cannot be ignored. This can add a double-edged effect on the learnt representations. On the one hand, representations may become more robust because small perturbations to $\mathbf{x}$ cannot easily affect classification results. On the other hand, Angular Gap is more likely to fall into local optimum because it takes more effort to update cosine similarities when feature norms are large. Therefore, in this work we propose global calibration and class-wise calibration to directly refine cosine similarities in the post-training.
\subsection{Calibration map}
\begin{figure}[htbp]
\centering
\subfigure[]{
\includegraphics[width=6.8cm]{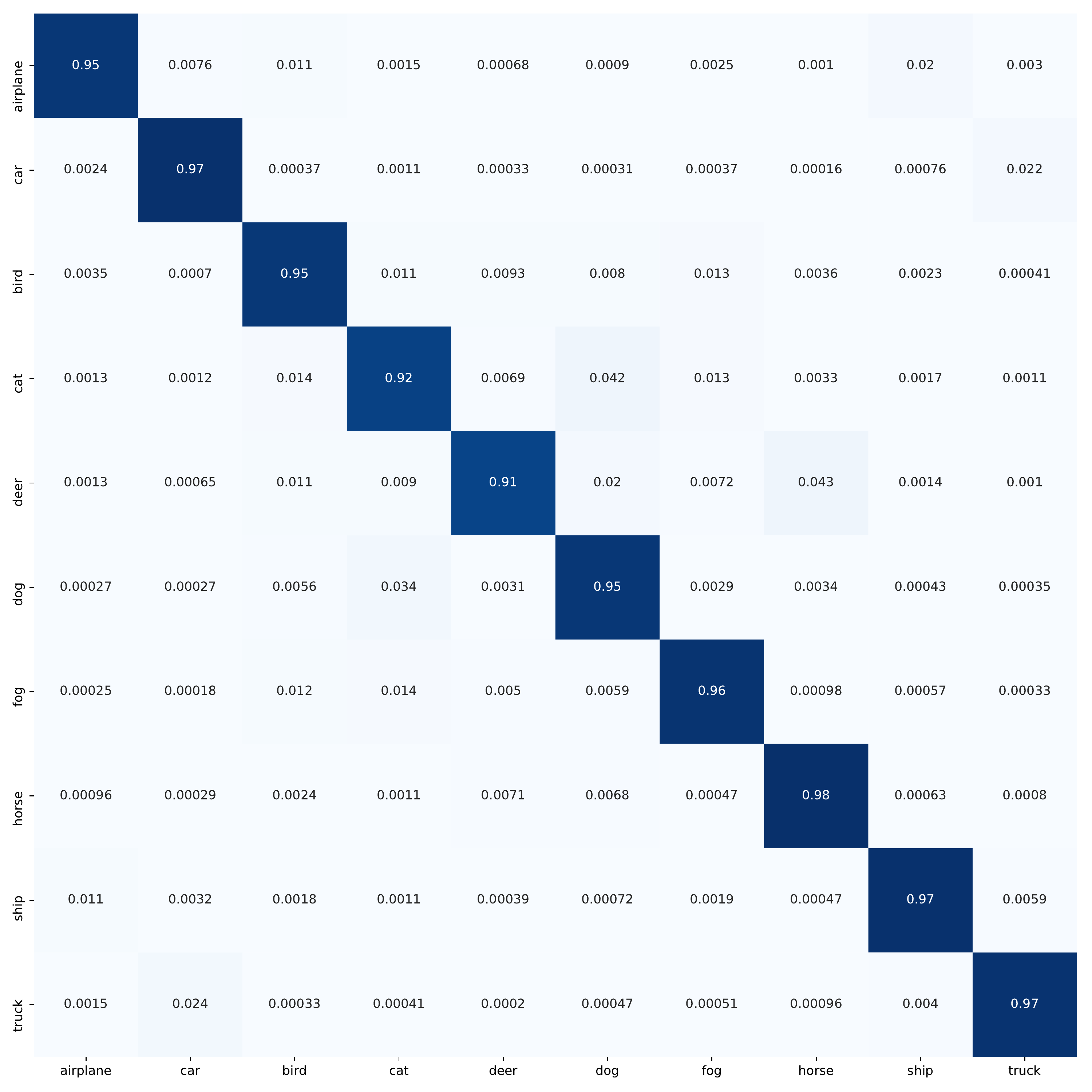}
}
\quad
\subfigure[]{
\includegraphics[width=6.8cm]{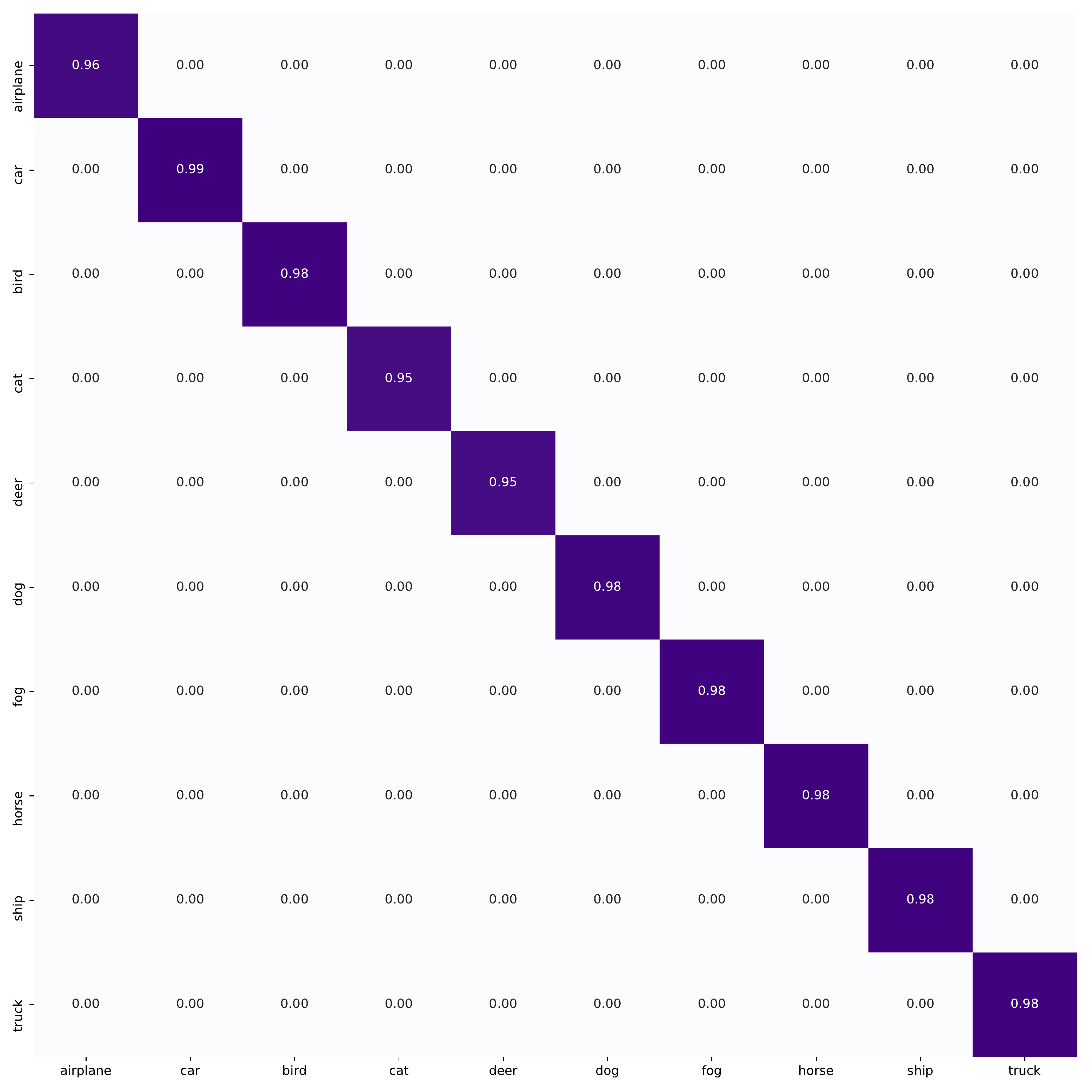}
}
\caption{Comparision of human predicted confusion matrix (a) and a class-wise calibration map (b).}
\label{fig:calibrationMap}
\end{figure}

Figure~\ref{fig:calibrationMap} (a) shows the confusion matrix of human classifiers reported by CIFAR10-H. The class level image difficulty can be represented by the precision of each class. Figure~\ref{fig:calibrationMap} (b) shows that class-wise calibration is able to capture class level difficulty in which the diagonal entries are learnt during the post-training and others are forced to be zeros.
\subsection{Reliability diagrams}
\begin{figure*}
\centering
\subfigure[Uncalibrated w/o FN]{
\includegraphics[width=0.62\linewidth]{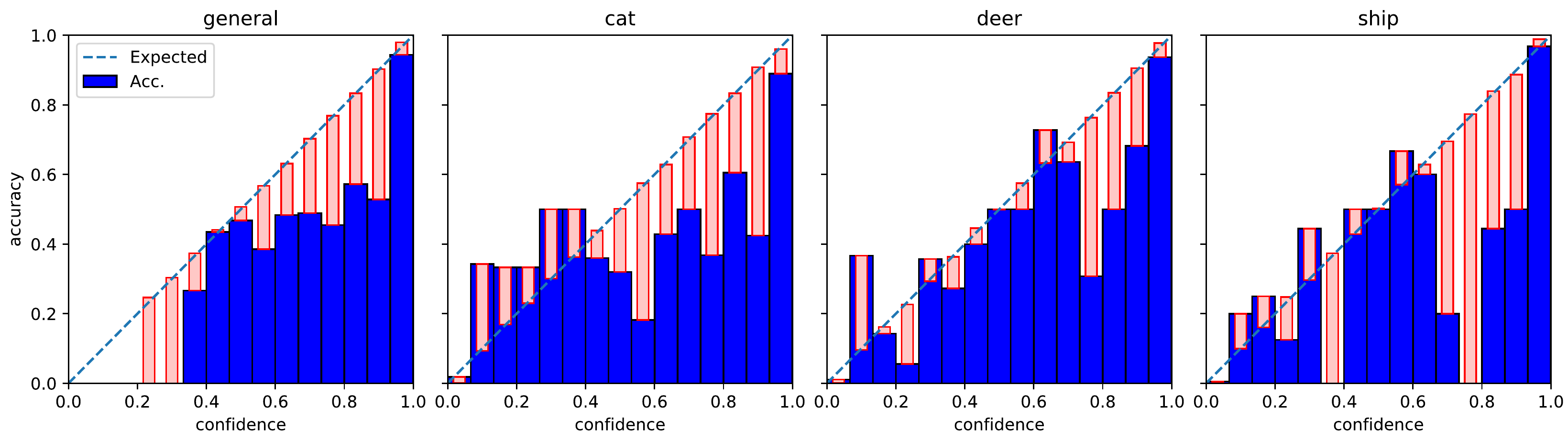}
}

\subfigure[Temperature scaling w/o FN]{
\includegraphics[width=0.62\linewidth]{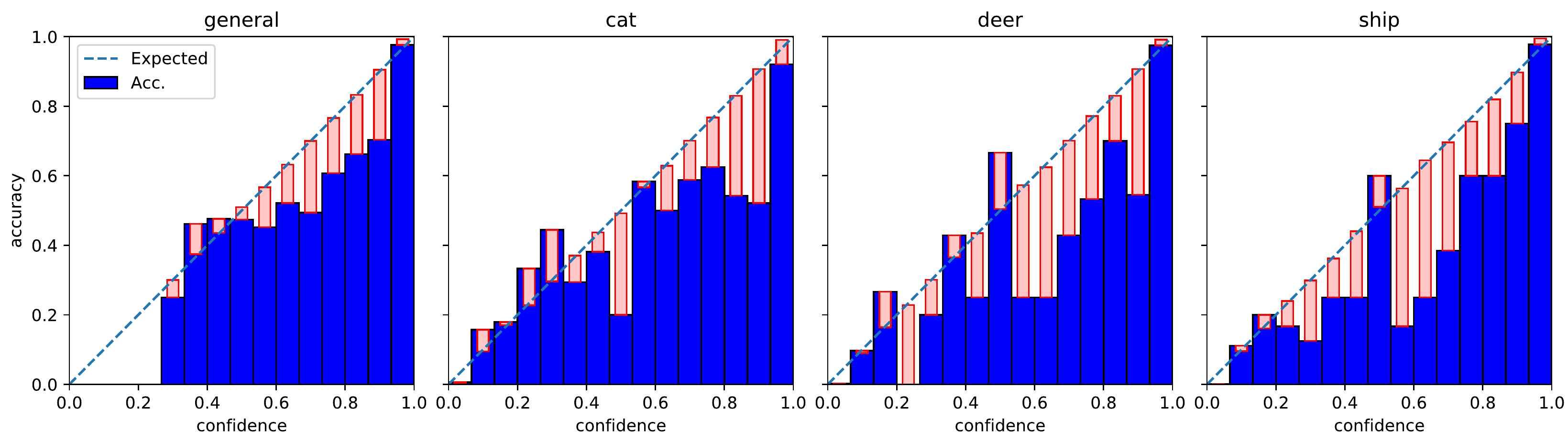}
}

    \subfigure[Uncalibrated with FN]{
\includegraphics[width=0.62\linewidth]{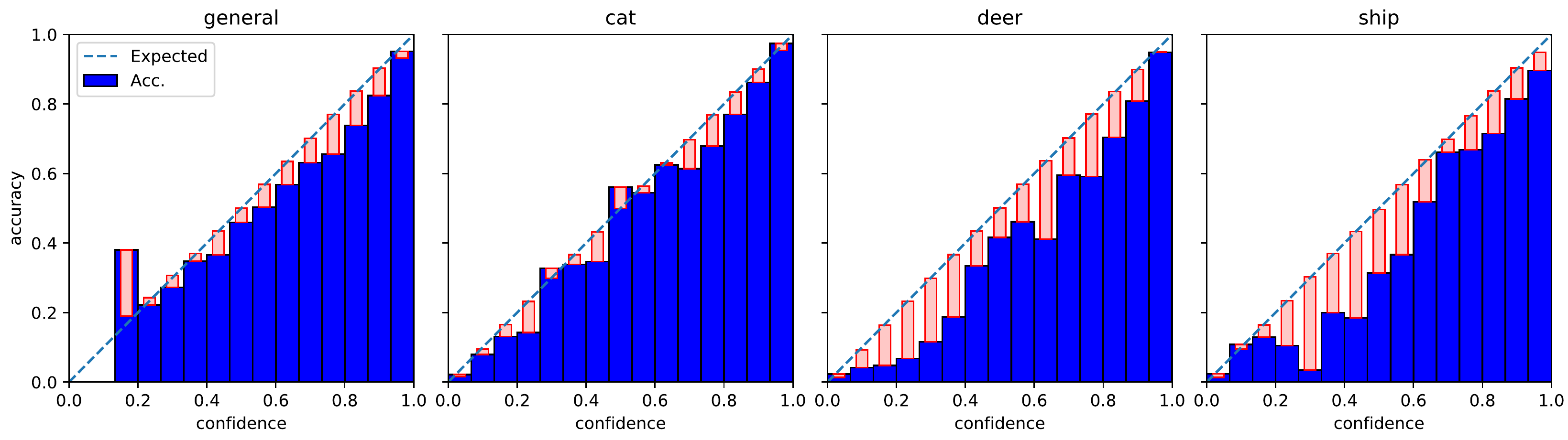}
}
\subfigure[Global calibration with FN]{
\includegraphics[width=0.62\linewidth]{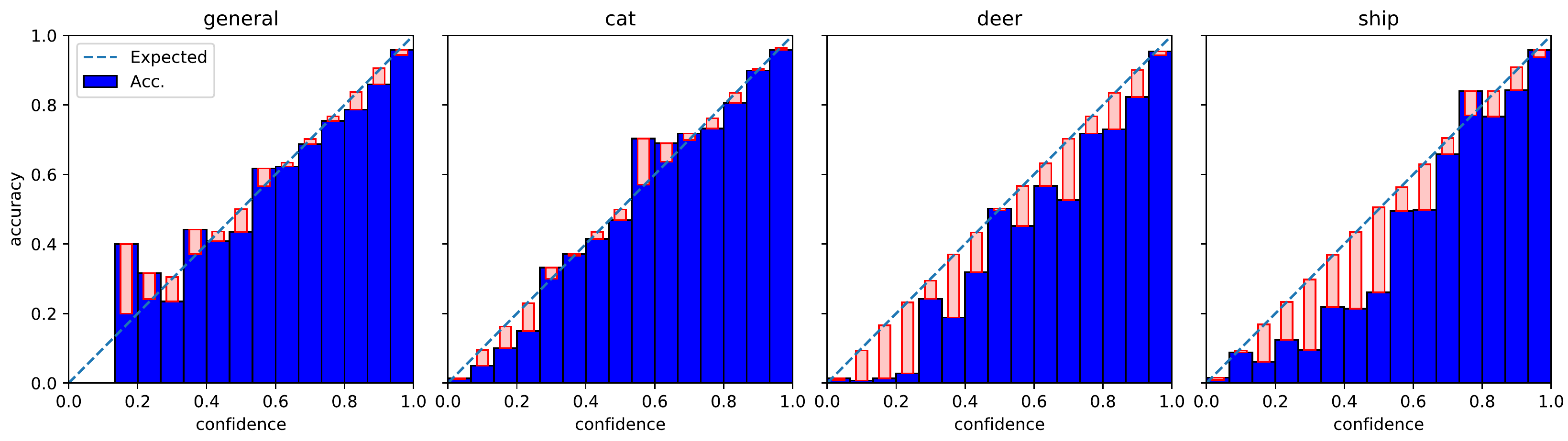}
}
\subfigure[Class-wise calibration with FN]{
\includegraphics[width=0.62\linewidth]{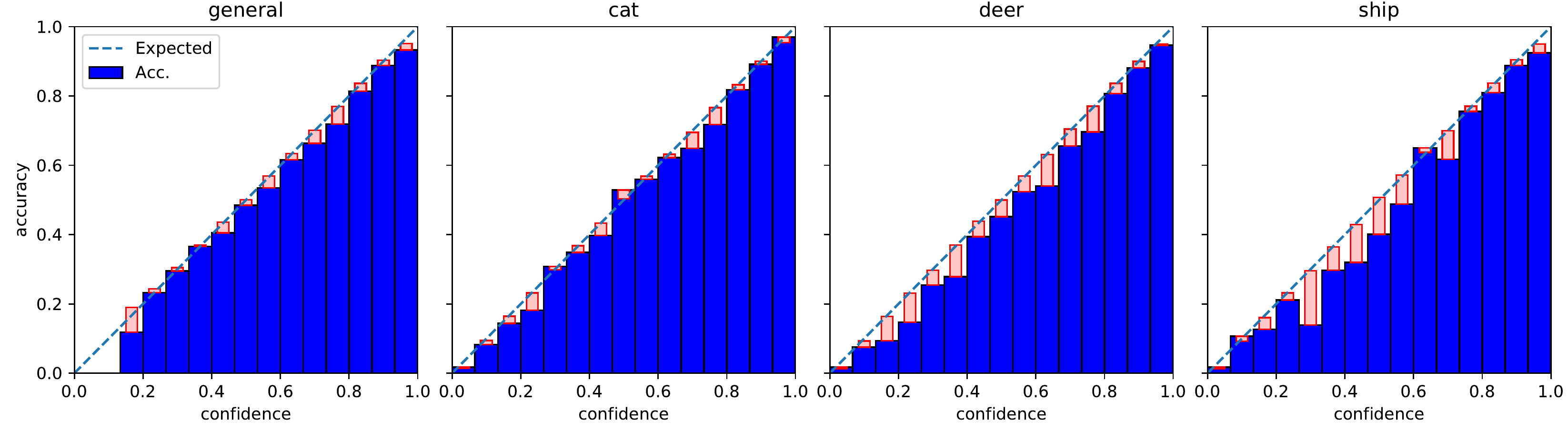}
}
    \caption{General reliability diagrams in the first column and class-wise reliability diagrams in the remaining columns. The diagrams visualize the calibration of ResNet18 models by comparing predictive confidence with observed accuracy of CIFAR10-H samples. Red bars indicate the gaps between expected accuracy (dash lines) and observed accuracy (blue bars) of the current confidence bin. (a) and (b) are pre-trained with cross-entropy loss and without feature normalization(FN). (b) uses temperature scaling to calibrate confidence. (c), (d) and (e) are trained with NSL. (d) and (e) applies the proposed global and class-wise calibration during the post-training respectively. Three out of ten classes are shown for visual clarification.}
    \label{fig:reliability}
\end{figure*}

\begin{figure*}
\centering
\subfigure[Early stopping]{
\includegraphics[width=0.62\linewidth]{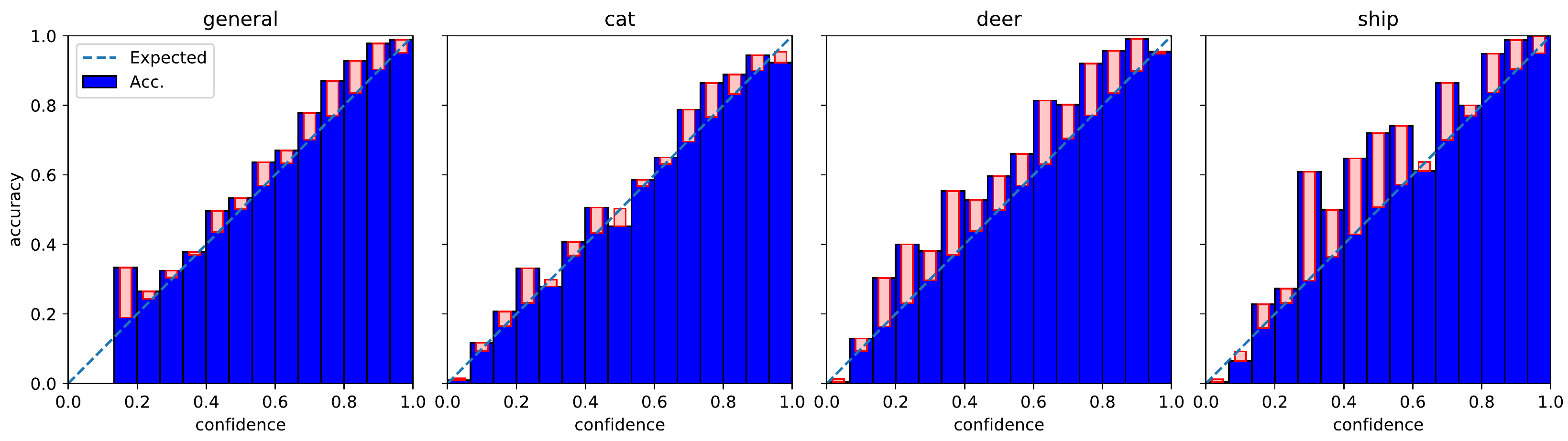}
}
\subfigure[Label smoothing]{
\includegraphics[width=0.62\linewidth]{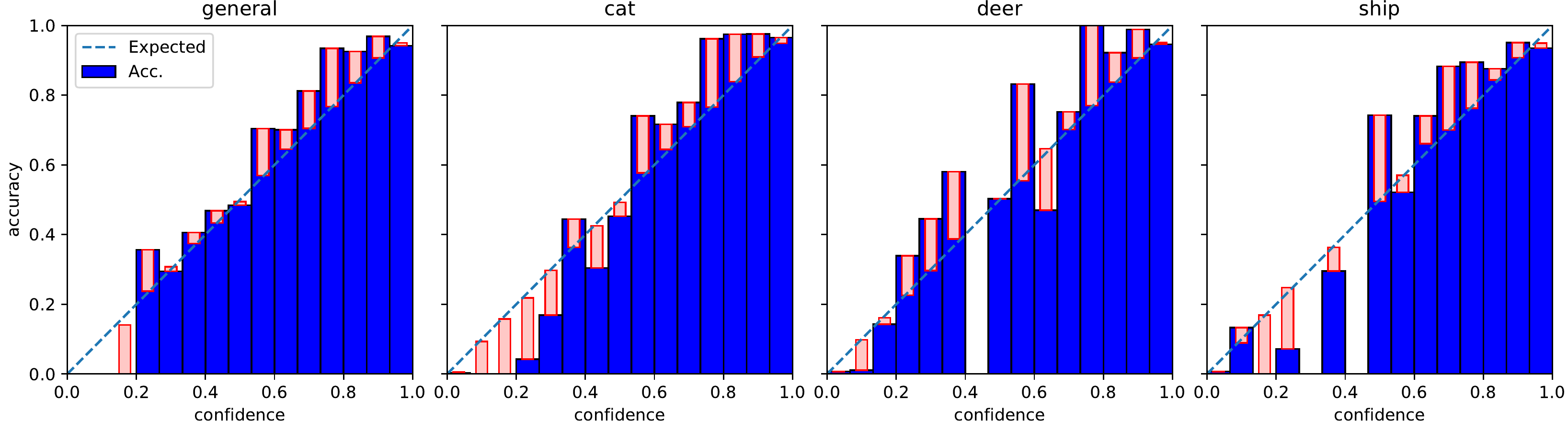}
}
\caption{Additional reliability diagrams of ResNet18 models trained with early stopping (a) and label smoothing (b). The first column shows the calibration results of the most probable softmax predictions, while the remaining columns show the calibration results of class-wise softmax predictions.}
\label{fig:others}
\end{figure*}

As visual representations of model calibration, reliability diagrams plot sample accuracy as a function of confidence.
Figure ~\ref{fig:reliability} shows reliability diagrams on the original difficulty estimators (a), and results of applying temperature scaling (b), feature normalization (c), global calibration (d) and class-wise calibration (e). These diagrams also explain the uncertainty of Angular Gap in the sense that Angular Gap can be nonlinearly mapped as confidence in hyperspherical learning. The first column shows the calibration results of the most probable softmax predictions, while the remaining columns show the calibration results of class-wise softmax predictions. The red bars visualize the gaps between expected accura With sample accuracy less than expected, the original model shows overconfidence to its predictions. After temperature scaling, the max output probabilities are calibrated but the gaps of the class-wise reliability diagrams are still obvious, which indicates considerable uncertainty exists in the sample confidence of less possible classes. Compared with the previous models trained with cross-entropy shown, models (c), (d) and (e) show less over-confidence, indicating feature normalization is able to reduce the uncertainty of example difficulty of CIFAR10-H samples. With a single learnable parameter, global calibration (d) is not enough to give well-calibrated confidence. Figure~\ref{fig:reliability} (e) shows that class-wise calibration is able to better calibrate sample confidence on all classes. Figure ~\ref{fig:others} (a) shows the calibration results of early stopping applied when the negative log likelihood stagnates. The model has become underconfident. Figure ~\ref{fig:others} (
b) shows the results of label smoothing (0.1), and class-wise reliability diagrams reveal its instability.
\begin{figure}[]
\centering
\includegraphics[width=\linewidth]{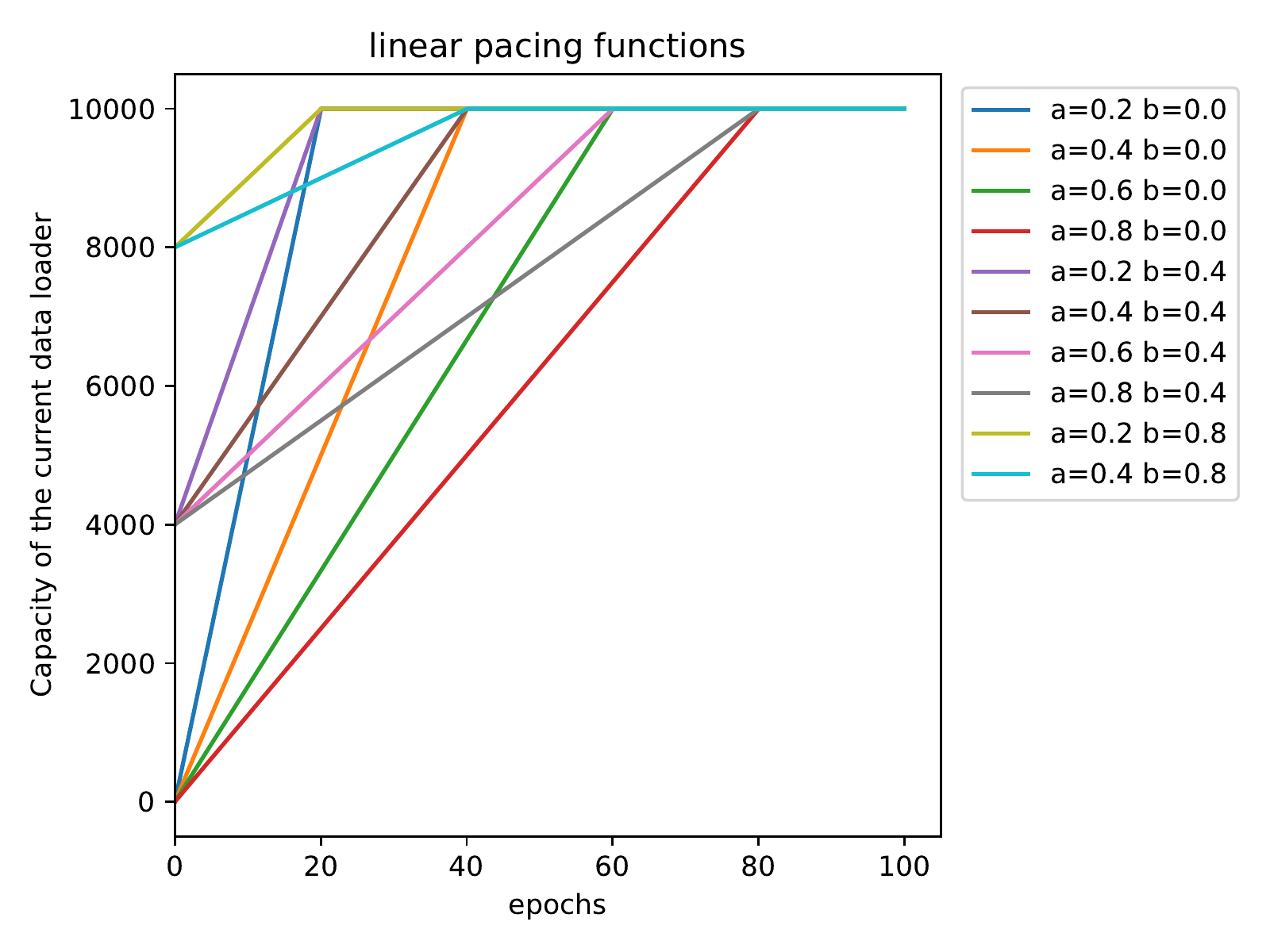}
\caption{Linear pacing functions for curriculum learning}
\label{fig:linear_pacings}
\end{figure}
\section{Standard curriculum learning}
\begin{figure*}
\centering
\includegraphics[width=\linewidth]{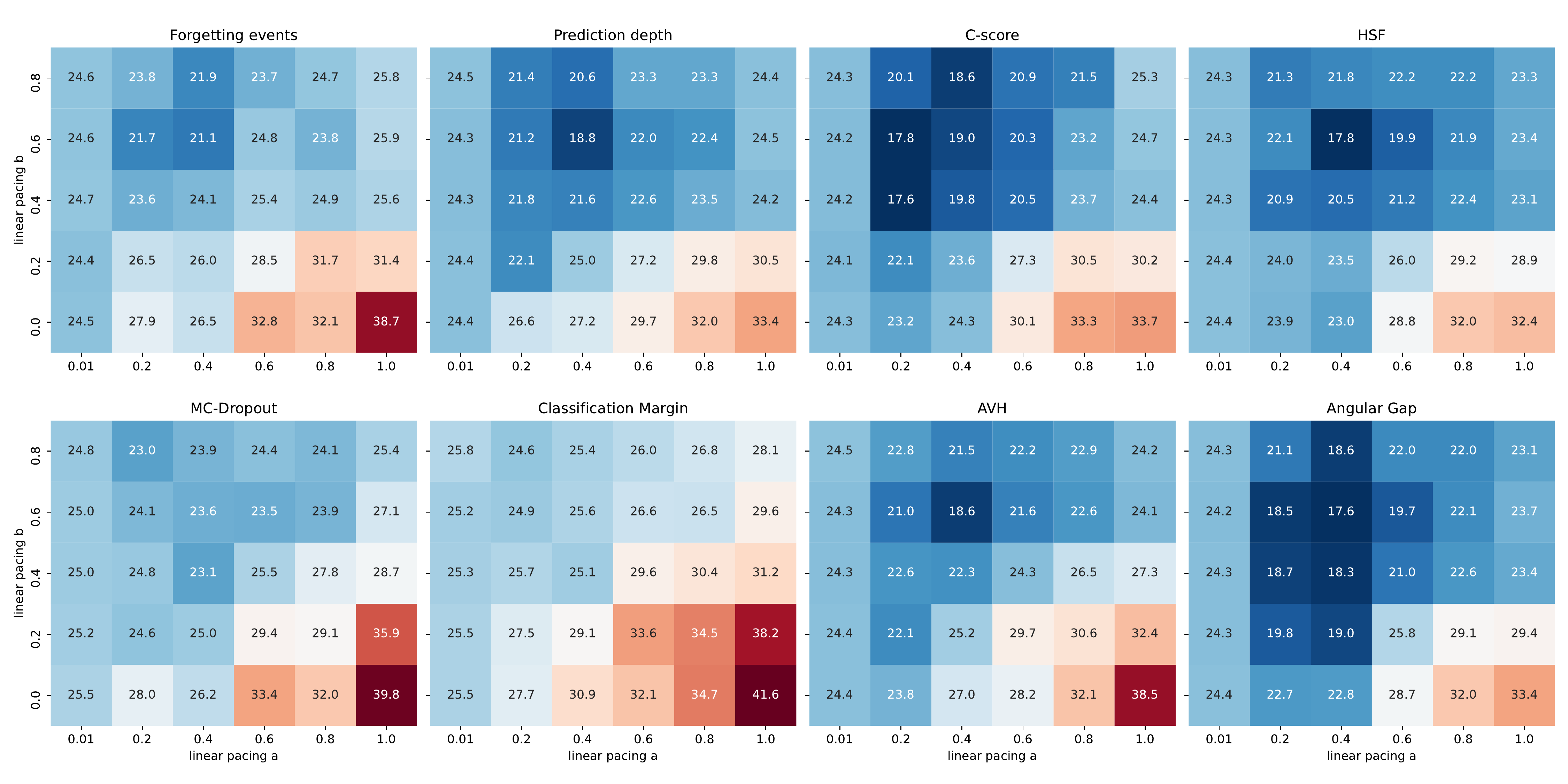}
\caption{Test errot rate (\%) of different difficulty metrics applied to standard curriculum learning on CIFAR10-H. Each difficulty metric corresponds to a heat-map for the parameter $a\in\{0.01, 0.2, 0.4, 0.6, 0.8, 1.0\}$ and the parameter $b\in\{0.0, 0.2, 0.4, 0.6, 0.8\}$. In each heat-map, a cell represents the median accuracy over five runs. Results in the first row show accuracy of image difficulty predicted by ensemble methods: (a) Forgetting events, (b) Prediction depth, (c) C-score, (d) HSF. The second row reports accuracy of image difficulty predicted by single-perspective methods: (e) MC-dropoout, (f) Classification margin, (g) AVH, (h) Angular Gap.}
\label{fig:grid}
\end{figure*}

In this section, we give more details about the standard curriculum learning evaluation that compares image difficulty metrics head-to-head mentioned in Section~\ref{sec:imp}. Standard curriculum learning, or Paced learning, designs a simple yet flexible curriculum with precomputed difficulty scores. This curriculum learning scheme contains two main steps. In the first step, data samples are sorted with precomputed difficulty scores in a fixed order. In the second step, a pacing function loads in a scheduled proportion of hard samples every epoch. We opt to use fixed training orders and linear pacing functions to simplify the evaluation. The linear pacing functions can be formally written as, 
\begin{equation}
    g_{(a,b)}(t) = N\frac{1-b}{aT}t + Nb,
\end{equation}
where the pacing functions start with $b$ percentage of training data and gradually add in samples until $a$ percentage of total iterations when the entire training set is fed, before the training continuing to the end. Figure~\ref{fig:linear_pacings} shows some examples of linear pacing functions for CIFAR10-H. For example, when $a$ is 0.2 and $b$ is 0.8, the linear pacing function corresponds to the olive line on the upper left.\\
The results for standard curriculum learning evaluation are shown in Figure \ref{fig:grid}. Angular Gap outperforms other image difficulty baselines of single-perspective methods, and is on par with the best performing ensemble method, C-score. The error rates of the upper left part of the heat-maps are generally lower than others. This shows that, in order to perform well, the model needs complex enough training materials during the early stage of training. 


\end{document}